\documentclass{article}
\pdfoutput=1
\usepackage{graphicx}
\usepackage{algorithm}
\usepackage{algorithmic}
\usepackage{tabularx} 
\usepackage{newfloat}
\usepackage{listings}
\usepackage{amssymb}
\usepackage{diagbox}
\graphicspath{{./}}
\usepackage{doi}

\usepackage{graphicx}
\usepackage{amsmath}
\usepackage{amssymb}
\usepackage{booktabs}

\usepackage{ragged2e} 

\newcommand{\HRule}[1]{\rule{\linewidth}{#1}}

\begin{document}
	
\title{
	\HRule{2pt} \\
	\textbf{Decoupling Deep Learning for Interpretable Image Recognition}
	\HRule{0.5pt}
	}

\author{
	\textbf{Yitao Peng},
	\textbf{Yihang Liu},
	\textbf{Longzhen Yang},
	\textbf{Lianghua He\textsuperscript{*}}\\
	College of Electronic and Information Engineering Tongji University\\
	4800 Cao’an Highway, Shanghai, China 201804\\
	\{pyt, 2111131, yanglongzhen, helianghua\}@tongji.edu.cn
}
\date{}

\maketitle	

\begin{abstract}	
	The interpretability of neural networks has recently received extensive attention. Previous prototype-based explainable networks involved prototype activation in both reasoning and interpretation processes, requiring specific explainable structures for the prototype, thus making the network less accurate as it gains interpretability. Therefore, the decoupling prototypical network (DProtoNet) was proposed to avoid this problem. This new model contains encoder, inference, and interpretation modules. As regards the encoder module, unrestricted feature masks were presented to generate expressive features and prototypes. Regarding the inference module, a multi-image prototype learning method was introduced to update prototypes so that the network can learn generalized prototypes. Finally, concerning the interpretation module, a multiple dynamic masks (MDM) decoder was suggested to explain the neural network, which generates heatmaps using the consistent activation of the original image and mask image at the detection nodes of the network. It decouples the inference and interpretation modules of a prototype-based network by avoiding the use of prototype activation to explain the network's decisions in order to simultaneously improve the accuracy and interpretability of the neural network. The multiple public general and medical datasets were tested, and the results confirmed that our method could achieve a 5\% improvement in accuracy and state-of-the-art interpretability compared with previous methods.
\end{abstract}

\thispagestyle{empty}
\newpage
\setcounter{page}{2}

\section{Introduction}
\label{sec:intro}

With the continuous development of neural networks (NNs) \cite{simonyan2014very,szegedy2015going, huang2017densely, liu2021swin, lee2022mpvit,liu2022convnet}, their interpretability is a research direction that has received extensive attention. It is challenging to make NNs have simultaneously good classification performance and interpretability. A large number of interpretability methods have been proposed in this regard. 

Saliency maps \cite{bach2015pixel, zhou2016learning,sundararajan2017axiomatic, chattopadhay2018grad,wang2020score,selvaraju2017grad} use localization as an explanation for predictions, but this only provides the network's area of interest for a given image, which does not fully represent the way the network makes its decisions \cite{rudin2019stop}. They lack generality and are not easily transferable to NNs with non-convolutional architecture.

\begin{figure}[t]
	\centering
	\includegraphics[width=0.9\linewidth]{{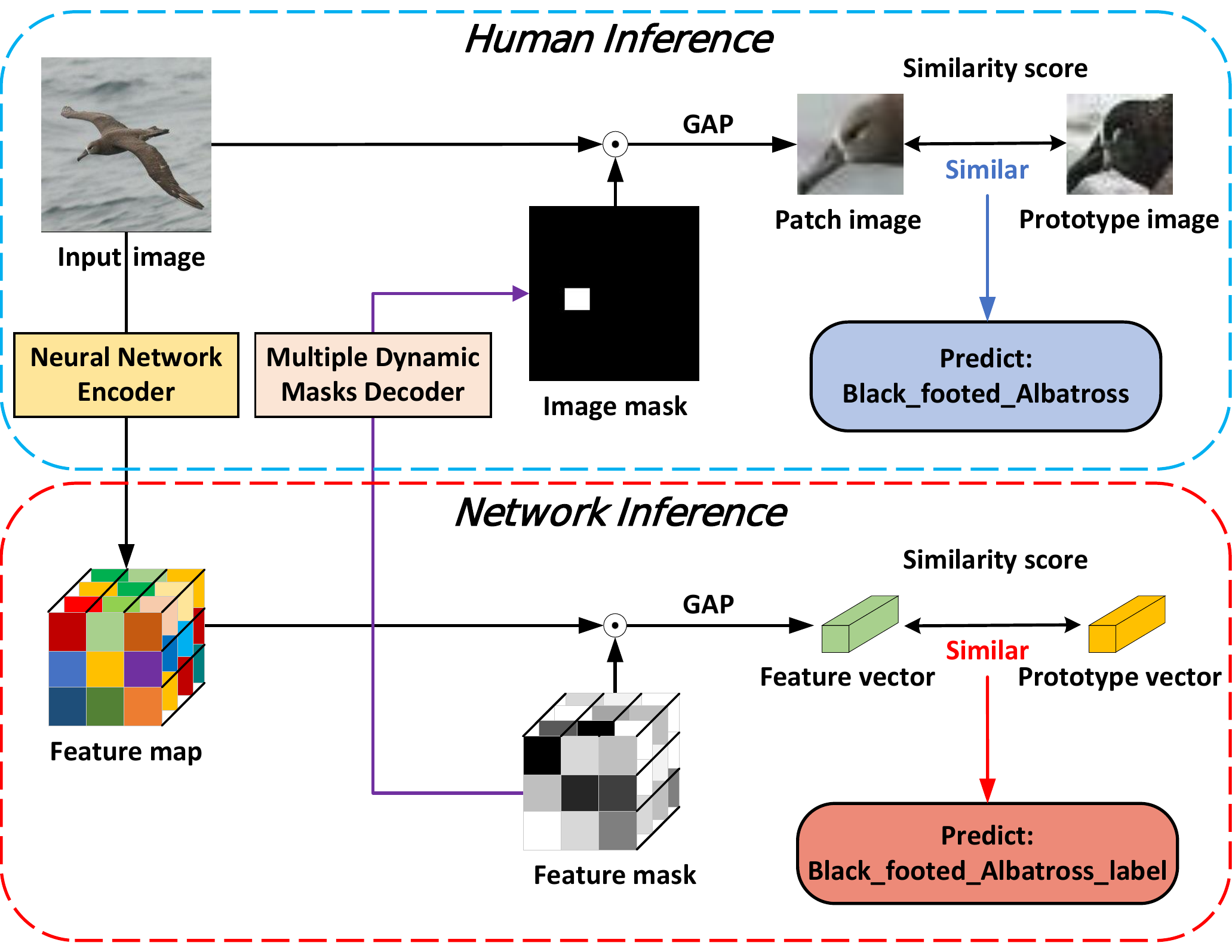}}
	
	\caption{The decision-making process of human and DProtoNet. Our approach enables the network to simulate the human reasoning process and transform the reasoning into human-comprehensible information. DProtoNet tries to keep the original structure of the backbone network, which is not unconstrained.}
	\label{fig:fig1}
\end{figure}

Interpretable models \cite{chen2019looks,singh2021these, singh2021interpretable, kim2021xprotonet, hase2019interpretable, li2018deep} are designed to function in a human-comprehensible way \cite{rudin2019stop}. They enable the network to learn feature templates for each class in the dataset, called prototypes. They predict the corresponding class by finding prototypes that are similar to the class. ProtoPNet \cite{chen2019looks}, Gen-ProtoPNet \cite{singh2021interpretable}, and XProtoNet \cite{kim2021xprotonet} use patches of different sizes in feature maps as prototypes for classification. However, none of these methods fully extract the information from the feature map. To make the prototype extracted by the network interpretable, they set a specific prototype structure, making the network subject to spatial constraints, thus leading to the reduction of network accuracy. These prototype-based networks \cite{chen2019looks, singh2021these, singh2021interpretable, kim2021xprotonet} think ``The patch of an input image that corresponds to the prototype should be the one that the prototype activates the most strongly on'' \cite{chen2019looks}. They localize prototypes and decision regions by upsampling feature maps with similarity activation maps produced by prototypes and then look for high activation regions as class activation maps (CAM) \cite{zhou2016learning} in the upsampled images. There is no complete theory to support that the activation area of the activation map can correspond to the decision area in the original image, thus the visualization generated by the previous method is not well interpretable and the positioning ability is inaccurate.

In this paper, a decoupling prototypical network (DProtoNet) is proposed to mine prototypes in data for interpretable classification. This network uses unrestricted feature masks to extract information in the feature map and relieve the constraints of the specific structure of the prototype on the latent space of the network. In addition, multi-image prototype learning is introduced to update the prototype by mixing the prototype features mined on multiple images so that the prototype can be represented as a distribution of certain types of features, avoiding the problem of introducing noisy prototypes when the image or network performance is low quality. By generalizing the extraction and learning of prototypes, DProtoNet enhances the expressive ability of prototypes and improves the accuracy of the network.

To solve the problems regarding the inaccurate localization of CAM and lack of theoretical support in prototype-based networks. A multiple dynamic masks (MDM) decoder was presented to visualize the decision regions of the network and provide mathematical proof. It is thought that when the network analyzes the image, only the decision region will promote the activation of the network at the specific node, and the region unrelated to the decision will not affect the activation of the network even if it is masked. Therefore, the MDM decoder sets detection nodes in the network and learns vectors through the consistent activation of the original image and the mask image on the detection nodes. It is noteworthy that the learning process of the MDM decoder conforms to public cognition so it is interpretable. The previous mask-based methods \cite{fong2017interpretable, chen2018learning, yuan2020interpreting} only perform activation consistent learning for masking the same size as the original image, which is prone to adversarial effects \cite{tu2020physically}. To reduce this, the MDM decoder stacks upsampled masks from multiple vectors of different sizes to generate the CAM. The mask generated by the MDM decoder can better preserve the spatial and semantic information of decision regions. Moreover, the prototype node in DProtoNet is set as the detection node, which can accurately locate the image information corresponding to the prototype and the decision region. MDM decoder does not take advantage of the internal architecture of the network thus it is generic.

As shown in Figure \ref{fig:fig1}, the DProtoNet keeps the prototype-based inference architecture to simulate the human inference process and uses the MDM decoder to explain the prototype and decision regions of DProtoNet. Furthermore, it decouples the inference and the interpretation module of the network, relieving the mutual constraints of accuracy and interpretability on network performance and improving the accuracy and interpretability of the network.

The key contributions of our work are as follows:

\begin{itemize}
	\item Unrestricted feature masks are proposed to mine global information from feature maps, thereby improving the expressiveness of features and prototypes.
	\item Multi-image prototype learning is introduced through which generalized prototypes can be learned.
	\item A general, interpretable, and powerful method, namely, the MDM decoder is presented for finding the basis for classification decisions in NNs, giving a mathematical proof of its feasibility.
	\item DProtoNet is proposed, which incorporates unrestricted feature masks, multi-image prototype learning, and the MDM decoder into encoder, inference, and interpretation modules, allowing the model to have good interpretability while improving accuracy.	
\end{itemize}

\section{Related Work}
\subsection{Saliency Maps}
Saliency methods produce a visual interpretation map that represents the importance of image pixels for network classification. Class activation mapping is a pioneering saliency method \cite{jalwana2021cameras}. \cite{zhou2016learning} uses global average pooling to integrate information from all features to obtain CAM. Nonetheless, CAM can only be used for specific mode structures. To address this limitation, Grad-CAM \cite{selvaraju2017grad} utilizes the gradient information of convolutional layers to obtain CAM. \cite{chattopadhay2018grad} proposed Grad-CAM++ to add an extra weight to measure the elements of the gradient map to precisely locate the CAM. To improve the versatility and accuracy of CAM, Score-CAM \cite{wang2020score} represents a gradient-free method for activation maps intuitively and understandably. Ablation-CAM \cite{ramaswamy2020ablation} analyzes the contribution of each factor to the network. These methods are various post-hoc attempts to interpret an already trained model and lack generality. Therefore, in this paper, a method is proposed to indicate the decision regions of the network, which can provide good interpretability for the network of any structure.

\begin{figure*}
	\centering
	\includegraphics[width=1.0\textwidth]{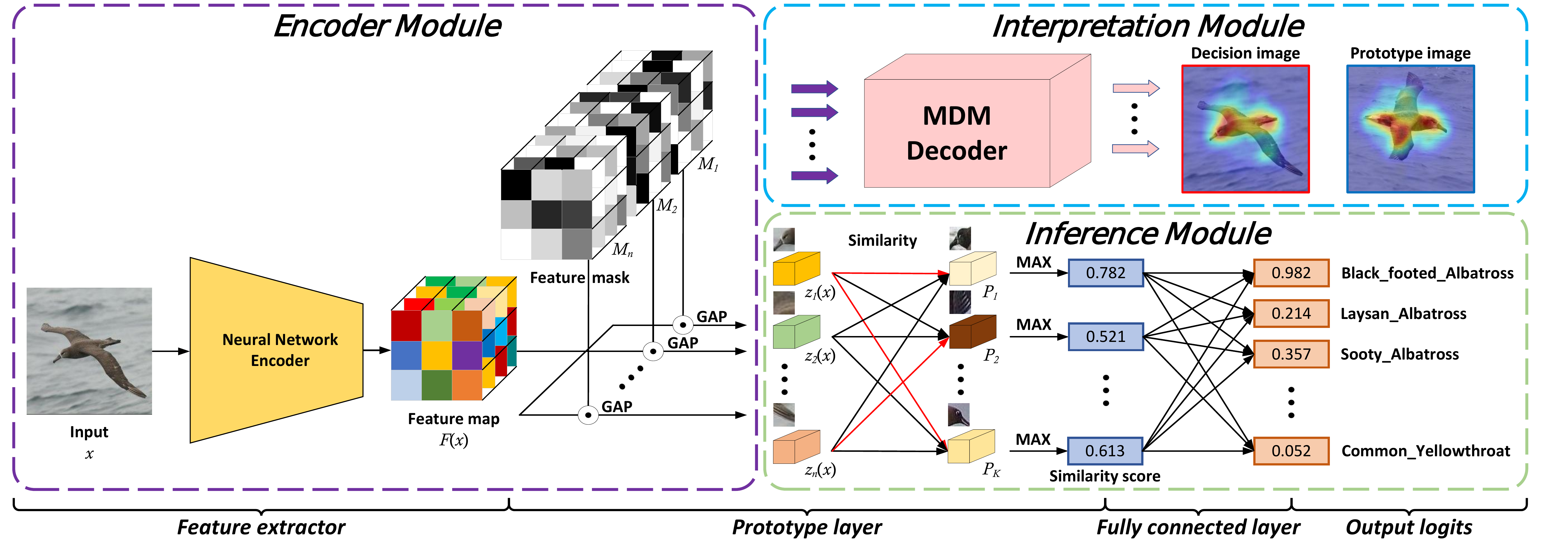}
	\hfill
	\caption{Overall architecture of DProtoNet. DProtoNet distinguishes image categories by comparing the features of an input image to the prototypes of each classification. It further generates decision and prototype images for reference through the MDM decoder.}
	\label{fig:fig2}
\end{figure*}

\subsection{Interpretable Models}
Setting the structure of the NN to mimic the human reasoning process makes the network interpretable. ProtoPNet \cite{chen2019looks} takes the $1 \times 1$ patches of feature maps as prototypes and uses them for classification. Additionally, NP-ProtoPNet \cite{singh2021these} fixes the last classification layer and exploits negative reasoning to improve the classification performance. To improve the adaptability of prototypes for different tasks, \cite{singh2021interpretable} proposed Gen-ProtoPNet, which improves the representation ability of the prototype by setting the prototype as the $h \times w$ patch on the feature map. Likewise, \cite{kim2021xprotonet} proposed XProtoNet, which sets the prototype as a feature vector with variable activation positions and sizes. These works set a specific shape for the prototype to limit the expressive ability of the network. Thus, a decoupled network architecture is set up in this study, which makes the network maintain the accuracy of the backbone network and have great interpretability.

\section{Methodology}
Figure \ref{fig:fig2} shows the overall architecture of our proposed framework, namely, DProtoNet, which consists of the feature extractor, prototype layer, fully connected layer, output logits, and MDM decoder. It can also be divided into encoder, inference, and interpretation modules. The inference and training of DProtoNet are described in Section \ref {Inference and Training of DProtoNet}. In addition, Sections \ref{Extraction of Prototype with Feature Mask} and \ref{Multi-image Prototype Learning} explain how to extract features within a global region and the prototype update method, respectively. Eventually, Section \ref{Multiple Dynamic Masks Decoder} introduces how to use the multiple dynamic masks (MDM) decoder to find the basis for decisions.

\subsection{Inference and Training of DProtoNet} \label{Inference and Training of DProtoNet}
\textbf{Classification Process.} Considering that DProtoNet has $K$ prototypes, input image $x \in R^{H\times W\times C}$, prototype $p_{j}$. The feature extractor is composed of a backbone network $f_{b}$ and a shaping network $f_{a}$. $f_{h}$ is the fully connected layer. Furthermore, $x$ passes $f_{b}$, $f_{a}$ to obtain feature map $F(x) \in R^{H_{1} \times W_{1} \times D_{1}}$, and then extract the feature vector $z_{i}(x)$. Similar to \cite{chen2019looks}, it calculates a similarity score between $z_{i}(x)$ and $p_{j}$, as well as activation $g_{p_{j}}(x)$ and logit $p(y^{c}|x)$.
\begin{equation} \label{eq1}
s(z_{i}(x), p_{j}) = ||z_{i}(x) - p_{j}||^{2}_{2}
\end{equation}
\begin{equation}
g_{p_{j}}(x) = g(F(x), p_{j}) = \ \mathop{max}\limits_{1 \leq i \leq n}\ log(\frac{s(z_{i}(x), p_{j}) + 1}{s(z_{i}(x), p_{j}) + \epsilon})
\end{equation}
\begin{equation}
p(y^{c}|x) = \sum_{j=1}^{K}w^{c}_{j}g(F(x), p_{j})
\end{equation}
where weight $w^{c}_{j}$ indicates how important each prototype $p_{j}$ is for the class $c$, $\epsilon$ prevents division by zero, and $n$ is the number of unrestricted feature masks.

\textbf{Training Scheme.} Training data is $\{(x_{i}, y_{i})\}^{n_{t}}_{i=1}$, which has $m$ classes. $Q_{k}$ represents the set of prototype $p_{j}$ belonging to class $k$, $w_{h}$ is the parameter of fully connected layer $f_{h}$, and $w_{h}^{(u,v)}$ is the ($u$,$v$)-th entry in $w_{h}$ that corresponds to the weight connection between the output of the $v$-th prototype unit $g_{p_{v}}$ and the logit of class $u$.
\begin{equation}
\begin{aligned}
L = & \frac{1}{n_{t}}\sum_{i=1}^{n_{t}}CrsEnt(f_{h} \circ g_{p} \circ F(x_{i}), y_{i})\\
&+\lambda_{1}Clst + \lambda_{2}Sep + \lambda_{3}l_{w_{h}} 
\end{aligned}
\end{equation}
where clustering cost minimization (Clst), separation cost minimization (Sep), and $l_{w_{h}}$ are defined as follows: 
\begin{equation}
Clst = \frac{1}{n_{t}}\sum_{i=1}^{n_{t}} \ \mathop{min}\limits_{j:p_{j}\in Q_{y_{i}}}\ \ \mathop{min}\limits_{k}\ ||z_{k}(x_{i})-p_{j}||^{2}_{2}
\end{equation}
\begin{equation}
Sep = - \frac{1}{n_{t}}\sum_{i=1}^{n_{t}} \ \mathop{min}\limits_{j:p_{j}\notin Q_{y_{i}}}\ \ \mathop{min}\limits_{k}\ ||z_{k}(x_{i})-p_{j}||^{2}_{2}
\end{equation}
\begin{equation}
l_{w_{h}}  = \sum_{u=1}^{m}\sum_{v:p_{v} \notin Q_{u}}|w^{(u,v)}_{h}|
\end{equation}

The cross-entropy loss penalizes misclassification. The Clst encourages each image to have some latent patchs that are at least close to a prototype of its own class. In addition, the Sep encourages each latent patch of the image to be far away from the prototypes not of its own class. By optimizing $l_{w_{h}}$, the prototypes that only belong to their own class participate in the classification. Similar to the training stage in ProtoPNet \cite{chen2019looks}, DProtoNet is trained by optimizing $L$.
\subsection{Extraction of Prototype with the Feature Mask} \label{Extraction of Prototype with Feature Mask}
\begin{figure}[t]
	\centering
	\includegraphics[width=1.0\linewidth]{{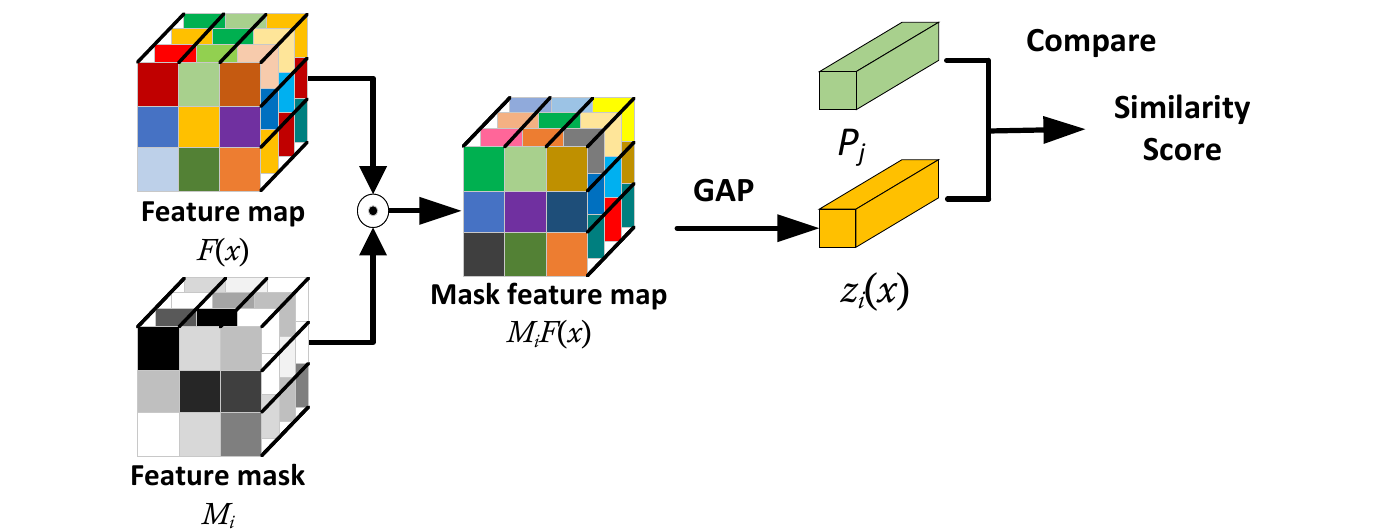}}
	
	\caption{Similarity score of the DProtoNet calculation process.}
	\label{fig:score}
\end{figure}

We randomly generate feature masks $\{M_{i}\}_{i=1}^{n}$, $M_{i} \in R^{H_{1} \times W_{1} \times D_{1}}$, each value in the elements of $M_{i} \in [0, 1]$. Equation (\ref{eq8}) generates feature vector $z_{i}$. Figure \ref{fig:score} displays the process, and GAP is global average pooling.
\begin{equation} \label{eq8}
z_{i}(x)=GAP(M_{i}F(x))
\end{equation}

The unrestricted mask $M_{i}$ is used to mine global information in the feature map. The set of prototypes generated by $M_{i}$ includes the set of prototypes generated by previous models \cite{chen2019looks,singh2021interpretable, kim2021xprotonet}. The expression ability of the prototype generated by $M_{i}$ is far greater than that of the previous models (refer to supplementary material for explanation).

Due to the arbitrariness of $M_{i}$, the information of feature map $F(x)$ is preserved to the greatest extent, relieving the spatial limitation of the prototype structure on the network and keeping the fitting ability of the backbone network unchanged. $M_{i}$ is versatile, thus we can generate prototypes with any custom number and style.

\subsection{Multi-image Prototype Learning} \label{Multi-image Prototype Learning}A multi-image prototype learning method is employed to update the prototype. Given that image $x_{i}$ belongs to class $k$, $\{x^{1}_{i}, x^{2}_{i}, ..., x^{R}_{i}\}$ is a group of images generated by $x_{i}$, which is the image after data augmentation. It is thought that the original data $x_{i}$ have the same characteristics as the data-augmented $x^{r}_{i}(r \in \{1, 2, ..., R\})$. We sum and project the patches most similar to the prototype $p_{j}$ in each $x^{r}_{i}$ as the update of $p_{j}$. Mathematically, the following update is performed for the prototype $p_{j}$ of class $k$ (i.e., $p_{j} \in Q_{k}$):
\begin{equation}
e_{r}  = \ \mathop{argmin}\limits_{e}\ ||z_{e}(x^{r}_{i})-p_{j}||_{2}
\end{equation}
\begin{equation} \label{eq10}
p_{j} \gets \ \mathop{argmin}\limits_{p}\ \sum_{r=1}^{R}||z_{e_{r}}(x^{r}_{i})-p||_{2}^{2}
\end{equation}

The $p_{j}$ generated by mixing multiple images is more robust than the  $p_{j}$ generated by a single image. From Equation (\ref{eq10}), according to the derivation, it can be known that the $p_{j}$ update formula is:
\begin{equation}
p_{j} = \frac{1}{R} \sum^{R}_{r=1}z_{e_{r}}(x^{r}_{i})
\end{equation}

\subsection{Multiple Dynamic Masks Decoder} \label{Multiple Dynamic Masks Decoder}
\begin{figure}[t]
	\centering
	\includegraphics[width=1.0\linewidth]{{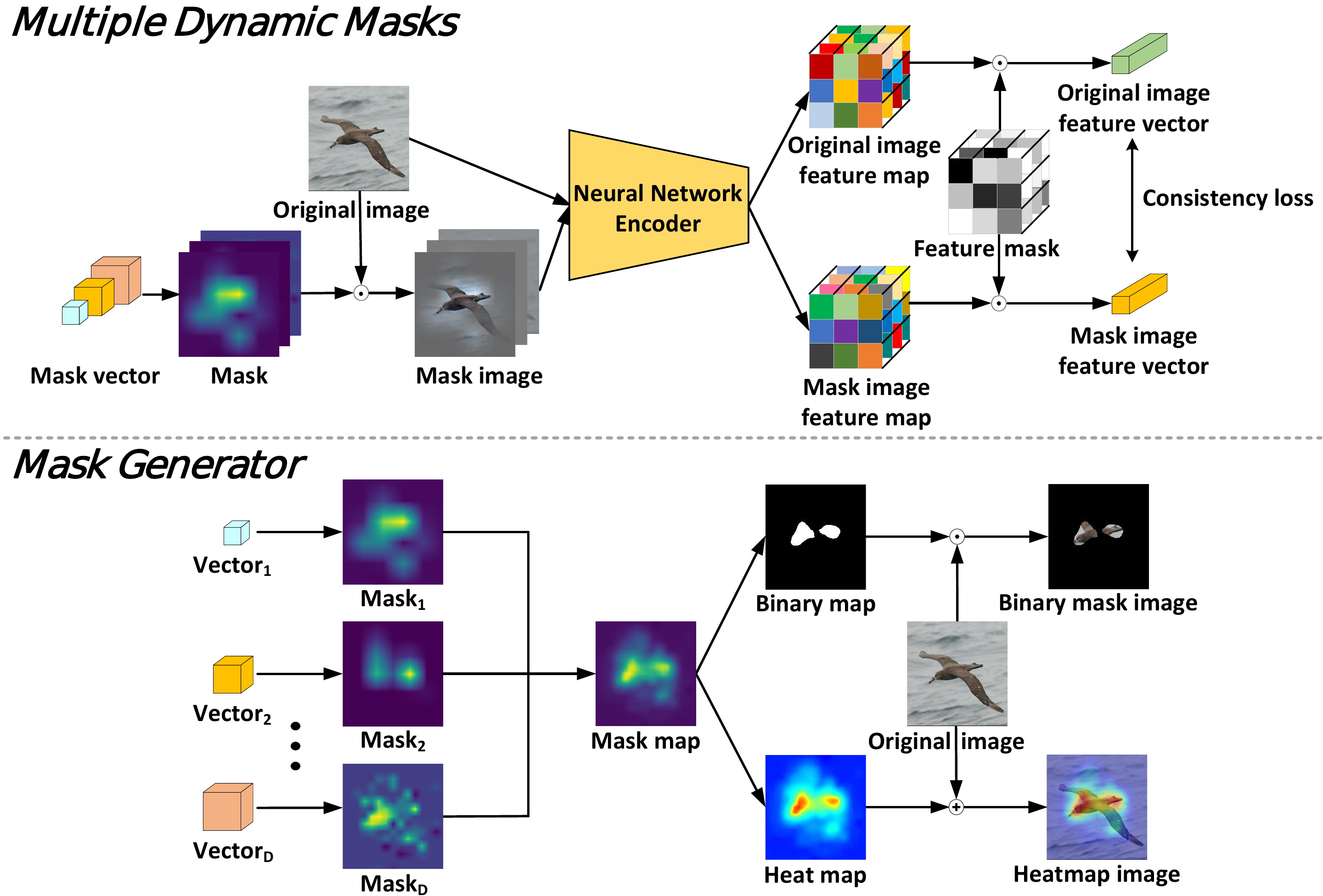}}
	
	\caption{
		The flow of a multiple dynamic masks decoder for generating saliency maps.}
	\label{fig:fig3}
\end{figure}

As depicted in Figure \ref{fig:fig3}, this decoder contains MDM and a mask generator. MDM learn the mask vectors of different sizes by constraining original and mask images to have consistent activation values at the detection nodes of the NN and mask vector values. A mask generator mixes the upsampled mask vectors to generate CAM in order to point out the decision regions of the NN. In DProtoNet, the prototype nodes in the network are chosen as detection nodes.

\textbf{Multiple Dynamic Masks.} The activation-consistent learning of the network is proposed through masks generated by multiple vectors of different sizes. The mask vector size is inversely proportional to its receptive field.

Mask vectors $\{d_{i}\}^{D}_{i=1}$, $d_{i} \in R^{a_{i} \times b_{i} \times 1}$, $d_{i}$ are initialized to a fixed value $\tau$. For any $i,j\in\{1,2,...,D\}$, if $i \neq j$ then $a_{i} \neq a_{j}$ or $b_{i} \neq b_{j}$. Upsample function $g(\cdot)$, $g(d_{i}) \in R^{H \times W \times 1}$, $d_{i}$ are upsampled to $g(d_{i})$ to mask the image.

Note that for input image $x$, DProtoNet classifies $x$ as $c$ and $p_{t}$ as a prototype belonging to the $c$ class. Further, $x^{p_{t}}$ is the image projected as $p_{t}$, and $M_{j_{p_{t}}}$ is the corresponding feature mask. Moreover, $M_{j_{x}}$ is the feature mask of $z_{j}(x)$ with the smallest similarity score to $p_{t}$. Additionally, $\{d^{x}_{i}\}^{D}_{i=1}$ and $\{d^{x^{p_{t}}}_{i}\}^{D}_{i=1}$ denote mask vectors generated based on $x$ and $x^{p_{t}}$, respectively.
\begin{equation}
j_{x}  = \ \mathop{argmin}\limits_{j}\ ||z_{j}(x)-p_{t}||_{2}
\end{equation}
\begin{equation}
j_{p_{t}}  = \ \mathop{argmin}\limits_{j}\ ||z_{j}(x^{p_{t}})-p_{t}||_{2}
\end{equation}

As show in Figure \ref{fig:fig3}, we train $\{d^{x}_{i}\}^{D}_{i=1}$, $\{d^{x^{p_{t}}}_{i}\}^{D}_{i=1}$ by the activation consistency between the mask image and the original image. Train $d^{x}_{i}$, $d^{x^{p_{t}}}_{i}$ by minimizing $L^{x}_{i}$, $L^{x^{p_{t}}}_{i}$.
\begin{equation} \label{eq13}
L^{x}_{i} = s(z_{j_{x}}(g(d^{x}_{i})x), z_{j_{x}}(x)) + \eta_{i}\sum^{a_{i}}_{u=1}\sum^{b_{i}}_{v=1}\frac{|d^{x}_{iuv}|}{|a_{i}b_{i}|}
\end{equation}
\begin{equation} \label{eq14}
L^{x^{p_{t}}}_{i} = s(z_{j_{p_{t}}}(g(d^{x^{p_{t}}}_{i})x^{p_{t}}), p_{t}) + \eta_{i}\sum^{a_{i}}_{u=1}\sum^{b_{i}}_{v=1}\frac{|d^{x^{p_{t}}}_{iuv}|}{|a_{i}b_{i}|}
\end{equation}
where $\eta_{i}$ is a regularization factor, and $s$ refer to Equation (\ref{eq1}). The mask vector retains the attention information of the image decision through the above-mentioned optimizations.

\textbf{Mask Generation.}
The trained $\{d^{x}_{i}\}^{D}_{i=1}$ and $\{d^{x^{p_{t}}}_{i}\}^{D}_{i=1}$ are upsampled to the original image size and mixed to generate CAM. Let $A^{x}$ and $A^{x^{p_{t}}}$ are the CAMs of $z_{j_{x}}(x)$ and $p_{t}$ in $x$ and $x^{p_{t}}$.
\begin{equation}
A^{x} = N(\{\sum^{D}_{i=1}g(d^{x}_{i}) \geq \gamma \}\sum^{D}_{i=1}g(d^{x}_{i}))
\end{equation}
\begin{equation}
A^{x^{p_{t}}} = N(\{\sum^{D}_{i=1}g(d^{x^{p_{t}}}_{i}) \geq \gamma \}\sum^{D}_{i=1}g(d^{x^{p_{t}}}_{i}))
\end{equation}
where $\gamma$ is the threshold, and $\{\cdot\}$ represents a truth-valued function, which is 1 if true; otherwise, it equals 0. $N(X) = \frac{X-min(X)}{max(X)-min(X)}$ is the normalization function.

As depicted in Figure \ref{fig:fig3}, binary mask and heatmap images are generated by multiplying and stacking the CAM and the original image.
\begin{equation}
A^{x}_{h} = \alpha x + \beta A^{x}, A^{x^{p_{t}}}_{h} = \alpha x^{p_{t}} + \beta A^{x^{p_{t}}}
\end{equation}
\begin{equation}
A^{x}_{b} = A^{x}x, A^{x^{p_{t}}}_{b} = A^{x^{p_{t}}}x^{p_{t}}
\end{equation}
where $\alpha$, $\beta$ are hyperparameters for image blending. Likewise, $A^{x}_{h}$ and $A^{x^{p_{t}}}_{h}$ are the heatmap images of $x$ and $x^{p_{t}}$. Moreover, $A^{x}_{b}$ and $A^{x^{p_{t}}}_{b}$ indicate the binary mask images of $x$ and $x^{p_{t}}$. They demonstrate those of prototype-like features and the regions of the prototype in $x$ and $x^{p_{t}}$ images, implying the regions of interest for DProtoNet classification and those of the prototype images used for the reference. This allows people to understand the decision-making process of DProtoNet.

\textbf{Feasibility of Multiple Dynamic Masks.} Let: $z$ represents the region in image $x$, and $f_{p}(z)$ denotes the activation of the NN $f$ at $p$ when the data of the region $z$ is taken as an input. $I(z)=kf_{p}(z)$, where $k$ is a constant greater than zero, $I(z) \in [0,1]$. $I(z)$ is the amount of information that region $z$ contributes to the activation of NN $f$ at position $p$.

Equations (\ref{eq13}) and (\ref{eq14}) can be expressed as follows:
\begin{equation} \label{eq19}
L(m,z) = [f_{p}(z)-f_{p}(mz)]^2 + \eta m
\end{equation}
where $z$ is all the areas of $d_{i}$, and $m$ is the corresponding mask value on it, $m\in [0,1]$.

There are two public cognition. When the corresponding regions on the original image do not intersect, it is considered that information $I$ of the contribution of the two regions to activation $f_{p}$ is irrelevant. Additionally, the greater contribution of the investigation region to the activation implies a greater the contribution to the information increment. Mathematically, $z_{1}$ and $z_{2}$ are the two regions of $d_{i}$, $i \in \{1,2,...,N\}$, and $g$ is the upsampling function.

if $g(z_{1}) \cap g(z_{2})=\varnothing$, then 
\begin{equation} \label{eq20}
I(z_{1}+z_{2})=I(z_{1})+I(z_{2})
\end{equation}

if $I(z_{1})<I(z_{2})$, then
\begin{equation} \label{eq21}
0\leq \frac{\partial I(mz_{1})}{\partial m} < \frac{\partial I(mz_{2})}{\partial m}
\end{equation}

Let: $z_{1}$ and $z_{2}$ demonstrate any two disjoint regions of $d_{i}$; $m_{1}$, $m_{2}$ are the mask values on $z_{1}$, $z_{2}$. From Equations (\ref{eq20}) and (\ref{eq21}), the following Equation (\ref{eq22}) can be proved, when $L(m, z)$ in Equation (\ref{eq19}) achieves the minimum value (refer to supplementary material for proof details).
\begin{equation} \label{eq22}
(I(z_{1}) - I(z_{2}))(m_{1} - m_{2}) \geq 0
\end{equation}

As shown in Equation (\ref{eq22}), optimization Equation (\ref{eq19}) can make the mask satisfy: the higher the mask value of the region with higher decision contribution results in more retaining of image information. However, the lower mask value of the region with lower decision contribution leads to retaining less image information. In DProtoNet, the prototype node is selected as the activation position $p$ so that the mask can mine the region represented by the prototype. 

\section{Experiments}
\subsection{Datasets and Baselines}
\textbf{Datasets.} Experiments were conducted on four image recognition datasets, including two general (CUB-200-2011 \cite{wah2011caltech} and Stanford Cars \cite{krause20133d}) and two medical (iChallenge-PM \cite{fu2019palm} and RSNA pneumonia \cite{gabruseva2020deep}) image datasets, followed by comparing the accuracy of interpretable and backbone networks on the four above-mentioned datasets. In the test dataset of CUB-200-2011, 10 images were randomly selected for each class, constituting a total of 2000 images. Finally, the recognition \cite{wang2020score} and localization \cite{wang2020score} abilities of the CAM on these images were compared as well.

\textbf{Baselines.}
The interpretable NNs (ProtoPNet \cite{chen2019looks}, NP-ProtoPNet \cite{singh2021these}, Gen-ProtoPNet \cite{singh2021interpretable}, and XProtoNet \cite{kim2021xprotonet}) and non-interpretable backbone networks (ResNet50 \cite{he2016deep}, VGG19 \cite{simonyan2014very}, and DenseNet121 \cite{huang2017densely}) were used as baselines to compare their accuracy with our proposed model. We adopted the recent state-of-the-art saliency map methods (Grad-CAM \cite{selvaraju2017grad}, Grad-CAM++ \cite{chattopadhay2018grad}, Score-CAM \cite{wang2020score}, and Ablation-CAM \cite{ramaswamy2020ablation}) and interpretable NNs (ProtoPNet \cite{chen2019looks}, NP-ProtoPNet \cite{singh2021these}, Gen-ProtoPNet \cite{singh2021interpretable}, and XProtoNet \cite{kim2021xprotonet}) generated CAMs as baselines in comparison with CAMs generated by our model for localization and recognition performance.

\begin{table}[H]
	\centering
	\resizebox{0.8\columnwidth}{!}{
	\begin{tabular}{@{}lccc@{}}
		\toprule
		Method & ResNet50 & VGG19 & DenseNet121 \\
		\midrule
		ProtoPNet \cite{chen2019looks} & 78.1 & 76.3 & 80.4 \\
		NP-ProtoPNet \cite{singh2021these} & 71.3 & 75.6 & 76.2 \\
		Gen-ProtoPNet \cite{singh2021interpretable} & 76.5 & 76.2 & 78.4 \\
		XProtoNet \cite{kim2021xprotonet} & 79.2 & 77.2 & 80.8 \\
		DProtoNet(ours) & \textbf{80.9} & \textbf{77.9} & \textbf{81.3} \\
		\hline
		\multicolumn{4}{c}{\textbf{CUB-200-2011} $\upuparrows$, \  \textbf{Stanford Cars} $\downdownarrows$ (dataset)}\\
		\hline
		ProtoPNet \cite{chen2019looks} & 85.9 & 87.7 & 86.9 \\
		NP-ProtoPNet \cite{singh2021these} & 83.2 & 85.2 & 83.6 \\
		Gen-ProtoPNet \cite{singh2021interpretable} & 85.6 & 85.8 & 84.1 \\
		XProtoNet \cite{kim2021xprotonet} & 84.7 & 87.3 & 84.3 \\
		DProtoNet(ours) & \textbf{86.5} & \textbf{89.2} & \textbf{89.3} \\
		\bottomrule
	\end{tabular}}
	\caption{Comparison results on general datasets.}
	\label{tab:general_images_accuracy}
\end{table}

\begin{table}[H]
	\centering
	\resizebox{0.8\columnwidth}{!}{
	\begin{tabular}{@{}lccc@{}}
		\toprule
		Method & Dataset & Accuracy & Sensitivity \\
		\midrule
		ProtoPNet \cite{chen2019looks} & RSNA & 73.2 & 35.5 \\
		NP-ProtoPNet \cite{singh2021these} & RSNA & 76.4 & 28.1 \\
		Gen-ProtoPNet \cite{singh2021interpretable} & RSNA & 76.9 & 34.8 \\
		XProtoNet \cite{kim2021xprotonet} & RSNA & 77.1 & 45.6 \\
		DProtoNet(ours) & RSNA & \textbf{82.2} & \textbf{49.8} \\
		\hline
		\hline
		ProtoPNet \cite{chen2019looks} & iChallenge-PM & 98 & 18.5 \\
		NP-ProtoPNet \cite{singh2021these} & iChallenge-PM & 97.25 & 0.4 \\
		Gen-ProtoPNet \cite{singh2021interpretable} & iChallenge-PM & 97.5 & 3.5 \\
		XProtoNet \cite{kim2021xprotonet} & iChallenge-PM & 98.25 & 3.3 \\
		DProtoNet(ours) & iChallenge-PM & \textbf{98.5} & \textbf{19.7} \\
		\bottomrule
	\end{tabular}}
	\caption{Comparison results on medical datasets.}
	\label{tab:medical_images_accuracy}
\end{table}

\subsection{Evaluation}
The performance of the model on nine evaluation metrics was tested, including accuracy \cite{singh2021these}, dice coefficient (DICE) \cite{laradji2021weakly}, IOU \cite{laradji2021weakly}, PPV \cite{laradji2021weakly}, sensitivity \cite{laradji2021weakly}, average drop (AD) \cite{chattopadhay2018grad}, average increase (AI) \cite{chattopadhay2018grad}, deletion scores (D) \cite{petsiuk2021black}, and insertion scores (I) \cite{petsiuk2021black}.

It should be noted that $TP$, $TN$, $FP$, and $FN$ are true positive, true negative, false positive, and false negative, respectively \cite{singh2021these}. DICE = $ \frac{2TP}{FP+2TP+FN} $, IOU = $ \frac{TP}{FP+TP+FN}$, PPV = $ \frac{TP}{TP+FP}$, sensitivity = $ \frac{TP}{TP+FN}$ and accuracy = $ \frac{number \ of \ correct \ predictions}{total \ number \ of \ cases}$ = $ \frac{TP+TN}{TP+TN+FP+FN}$. AD = $ \sum_{i=1}^{N} \frac{100\max (0,Y^{c}_{i}-O^{c}_{i})}{Y^{c}_{i}}$, AI = $ \sum_{i=1}^{N}\frac{100Sign(Y^{c}_{i}<O^{c}_{i})}{N}$. $Y^{c}_{i}$ and $O^{c}_{i}$ denote the prediction score of class $c$ in the original image $i$ and explained map, respectively. Certain percentile pixels of the original image were removed to generate an explained map. $Sign(\cdot)$ is an indicator function, and it is 1 if true. D and I measures are the deletion and insertion of pixels from the original image in descending order of the CAM activation value, respectively, and generate the area under the probability curve described by the predicted probability result of the deleted or inserted image. 

The CAM as a 0-1 binary mask was generated according to a percentage threshold. Then, dice coefficient, IOU, PPV, and sensitivity with the segmentation foreground or bounding box of the image were calculated to measure the localization \cite{wang2020score} ability of the CAM. In addition, AD, AI, D, and I were used to measure the recognition \cite{wang2020score} ability of the CAM, and accuracy was employed to determine the classification performance of the model.

\begin{figure}[t]
	\centering
	\includegraphics[width=1.0\linewidth]{{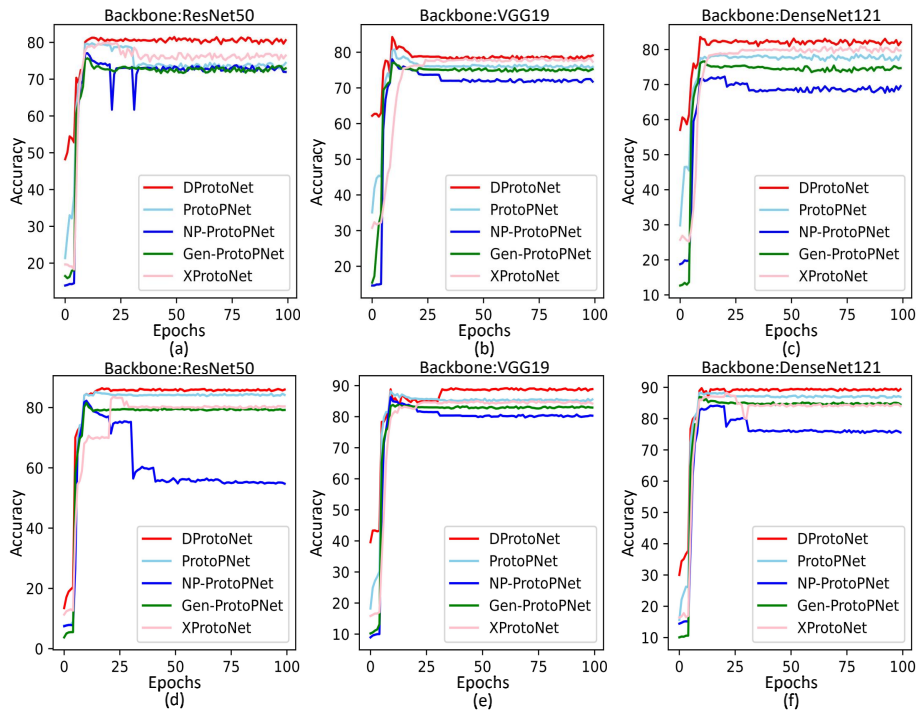}}
	
	\caption{Comparison of training stage accuracy.}
	\label{fig:accuracy}
\end{figure}

\subsection{Experimental Details}
Overall, 10 prototypes were considered for each class. Each image was rotated, perspectived, sheared, and distorted to generate augmented images. All the images were cropped to $224 \times 224$. A shaping network consists of two $1 \times 1$ convolutional layers with ReLU activation between them. Hyperparameters were derived using five-fold cross-validation, $\alpha=0.5$, $\beta=0.3$, $\gamma=3$, $\tau=0.5$, $\lambda_{1}=0.8$, $\lambda_{2}= -0.08$, $\lambda_{3}= 1e-4$, $\epsilon=1e-12$, $D=10$, $H_{1}=W_{1}=7$, $D_{1} = 512$, $a_{i} = b_{i} = 5 + i$, $\eta_{i}=10, \ i \in \{1, 2, ..., 10\}$. The number of feature masks was 720. The Adam optimizer was used, and the learning rates of backbone layer, shaping layer, prototype layer, and fully connected layer in the DProtoNet were set to $1e-4$, $3e-3$, $3e-3$, $1e-4$, respectively. The parameters of the backbone NNs were initialized to the values pre-trained on ImageNet \cite{deng2009imagenet}. The prototype channel and the batch were 512 and 60, respectively. Each mask vector was trained for 800 iterations. The initial stage is the first five epochs. Then, it was the joint stage, and the prototype update was performed every 10 epochs. In examining the sensitivity, the binary mask threshold was set to the top 50\% on the RSNA and iChallenge-PM datasets. In examining the DICE, IOU, PPV and sensitivity, the binary mask thresholds were set to the top 20\% on the CUB-200-2011.

\begin{table}[H]
	\centering
	\resizebox{0.8\columnwidth}{!}{
	\begin{tabular}{@{}lccc@{}}
		\toprule
		Backbone & Dataset & DProtoNet & Baseline \\
		\midrule
		ResNet50 & CUB-200-2011 & 80.9 & 81.2 \\
		VGG19 & CUB-200-2011 & 77.9 & 75.5 \\
		DenseNet121 & CUB-200-2011 & 81.3 & 80.6 \\
		ResNet50 & Stanford Cars & 86.5 & 86.3 \\
		VGG19 & Stanford Cars & 89.2 & 88.6 \\
		DenseNet121 & Stanford Cars & 89.3 & 89.8 \\
		ResNet50 & RSNA & 82.2 & 79.6 \\
		VGG19 & RSNA & 79.3 & 78.2 \\
		DenseNet121 & RNSA & 80.6 & 79.9 \\
		ResNet50 & iChallenge-PM & 98.5 & 98.75 \\
		VGG19 & iChallenge-PM & 98.25 & 98.5 \\
		DenseNet121 & iChallenge-PM & 98.75 & 98.5 \\
		\bottomrule
	\end{tabular}}
	\caption{Accuracy comparison of four above-mentioned datasets.}
	\label{tab:Backbone_Accuracy}
\end{table}

\begin{table}[H]
	\centering
	\resizebox{0.8\columnwidth}{!}{
	\begin{tabular}{@{}lcccc@{}}
		\toprule
		Backbone & M=1 & M=10 & M=20 & M=40 \\
		\midrule
		ResNet50 \cite{he2016deep} & 80.5 & 80.6 & 80.8 & \textbf{80.9} \\
		VGG19 \cite{simonyan2014very} & 77.5 & 77.7 & 77.8 & \textbf{77.9} \\
		DenseNet121 \cite{huang2017densely} & 80.9 & 81.1 & 81.2 & \textbf{81.3} \\
		\bottomrule
	\end{tabular}}
	\caption{Comparison of multi-image prototype learning.}
	\label{tab:multi-image}
\end{table}

\subsection{Network Classification Performance}
\textbf{Comparison with Interpretable Networks.} The accuracy of DProtoNet was compared with recent interpretable models. The findings (Table \ref{tab:general_images_accuracy}) revealed that DProtoNet has achieved state-of-the-art accuracy on ResNet50, VGG19, and DenseNet121 backbones on general datasets. Based on the data (Table \ref{tab:medical_images_accuracy}), the accuracy and sensitivity of each model were compared with ResNet50 as a backbone. On the RSNA dataset, the accuracy of DProtoNet was 5.1\% higher than that of the previous state-of-the-art model, and the sensitivity was 4.2\% higher. On the iChallenge-PM dataset, DProtoNet outperformed previous models in terms of both accuracy and sensitivity. DProtoNet had good accuracy for pathological images, and its decision regions could well localize real pathological regions. The reasoning process of DProtoNet complied with the process of “diagnosed disease based on pathological features found”, which is interpretable and can be recognized by clinicians. Figure \ref{fig:accuracy} illustrates the variation in accuracy for each model trained on the general datasets. Figures (a), (b), and (c), as well as (d), (e), and (f), are the results of the bird \cite{wah2011caltech} and car \cite{krause20133d} datasets, respectively. DProtoNet converged faster than the other models and achieved the best accuracy. All other interpretable networks suffered from accuracy degradation after performing the prototype update operation. After DProtoNet performs the prototype update, the classification performance of the network is almost not degraded, and the accuracy of the network is stable. This is because DProtoNet retains the interpretable inference process, but does not set a specific structure for the prototype so that the values of the network will not mutate after the update of the prototype.

\textbf{Comparison with Backbone Networks.} Table \ref{tab:Backbone_Accuracy} provides a comparison of the accuracy of DProtoNet and its backbone networks (ResNet50, VGG19, and DenseNet121) on four datasets. The accuracy of DProtoNet is almost comparable to the classification performance of the backbone network on some datasets, and the accuracy on the other datasets exceeds the accuracy of the backbone network. DProtoNet achieves interpretability without degrading its accuracy. In addition, it extracts the global information of the feature map by using the unrestricted feature mask with the strong expressive ability and retains the fitting ability of the backbone as much as possible so that the accuracy of DProtoNet can be comparable to that of the backbone network. Moreover, DProtoNet treats the introduced backbone network as a black box and does not modify the internal architecture of the encoder network, thus it can be widely applied to the existing networks.

\textbf{Evaluation of Multi-image Prototype Learning.} Table \ref{tab:multi-image} presents the accuracy of DProtoNet learned on the bird \cite{wah2011caltech} dataset with ResNet50, VGG19, and DenseNet121 as backbones mixed with different numbers of M images as prototypes. M = 1 indicates the single-image prototype learning method used by previous models \cite{chen2019looks, singh2021these, singh2021interpretable, kim2021xprotonet}. An increase in M leads to higher network accuracy. This method can generally learn prototypes, improving the accuracy of the prototype-based interpretable network.

\begin{figure}[H]
	\centering
	\includegraphics[width=0.9\linewidth]{{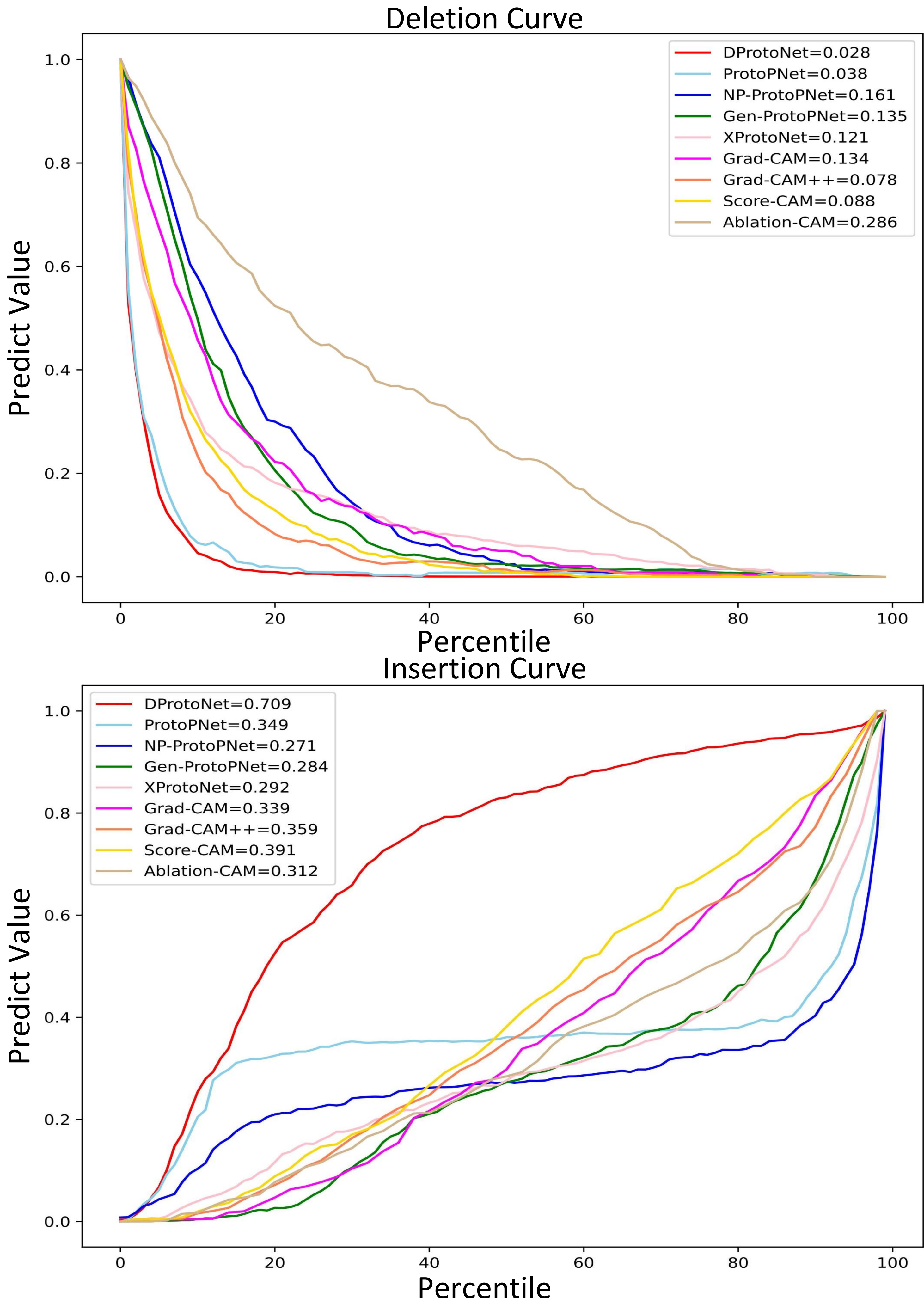}}
	
	\caption{Deletion and insertion curves of the above methods.}
	\label{fig:deletion_insertion}
\end{figure}

\begin{table}[H]
	\centering
	\resizebox{1\columnwidth}{!}{
	\begin{tabular}{@{}lcccccccc@{}}
		\toprule
		Method & AD & AI & D & I & DICE & IOU & PPV & Sensitivity \\
		\midrule
		Grad-CAM \cite{selvaraju2017grad} & 27.8 & 14.2 & 0.134 & 0.339 & 0.288 & 0.186 & 0.292 & 0.336 \\
		Grad-CAM++ \cite{chattopadhay2018grad} & 67.3 & 3.7 & 0.078 & 0.359 & 0.476 & 0.338 & 0.465 & 0.557 \\
		Score-CAM \cite{wang2020score} & 44.5 & 13.2 & 0.088 & 0.391 & 0.409 & 0.284 & 0.411 & 0.468 \\
		Ablation-CAM \cite{ramaswamy2020ablation} & 82.4 & 4.9 & 0.286 & 0.312 & 0.231 & 0.151 & 0.232 & 0.265 \\
		ProtoPNet \cite{chen2019looks} & 75.1 & 3.2 & 0.038 & 0.349 & 0.527 & 0.373 & 0.509 & 0.639 \\
		NP-ProtoPNet \cite{singh2021these} & 45.7 & 11.5 & 0.161 & 0.271 & 0.071 & 0.044 & 0.076 & 0.075 \\
		Gen-ProtoPNet \cite{singh2021interpretable} & 55.2 & 15.9 & 0.135 & 0.284 & 0.287 & 0.178 & 0.292 & 0.324 \\
		XProtoNet \cite{kim2021xprotonet} & 76.1 & 4.7 & 0.121 & 0.292 & 0.415 & 0.273 & 0.403 & 0.492 \\
		DProtoNet(ours) & \textbf{17.5} & \textbf{21.1} & \textbf{0.028}& \textbf{0.709} & \textbf{0.548} & \textbf{0.391} & \textbf{0.531}& \textbf{0.651} \\
		\bottomrule
		\multicolumn{9}{c}{\textbf{ResNet50} $\upuparrows$, \  \textbf{DenseNet121} $\downdownarrows$ (backbone)} \\
		\toprule
		Grad-CAM \cite{selvaraju2017grad} & 49.3 & 12.8 & 0.148 & 0.563 & 0.319 & 0.205 & 0.311 & 0.449 \\
		Grad-CAM++ \cite{chattopadhay2018grad} & 71.3 & 4.6 & 0.045 & 0.315 & 0.521 & 0.365 & 0.509 & 0.607 \\
		Score-CAM \cite{wang2020score} & 37.5 & 13.9 & 0.091 & 0.632 & 0.466 & 0.329 & 0.461 & 0.541 \\
		Ablation-CAM \cite{ramaswamy2020ablation} & 89.6 & 2.5 & 0.127 & 0.185 & 0.254 & 0.163 & 0.254 & 0.294 \\
		ProtoPNet \cite{chen2019looks} & 31.2 & 16.9 & 0.056 & 0.631 & 0.289 & 0.183 & 0.244 & 0.519 \\
		NP-ProtoPNet \cite{singh2021these} & 90.2 & 1.4 & 0.424 & 0.211 & 0.364 & 0.232 & 0.312 & 0.612 \\
		Gen-ProtoPNet \cite{singh2021interpretable} & 60.2 & 11.9 & 0.161 & 0.261 & 0.298 & 0.186 & 0.298 & 0.342 \\
		XProtoNet \cite{kim2021xprotonet} & 31.8 & 17.3 & 0.102 & 0.617 & 0.397 & 0.256 & 0.389 & 0.473 \\
		DProtoNet(ours) & \textbf{15.2} & \textbf{19.8} & \textbf{0.041}& \textbf{0.745} & \textbf{0.626} & \textbf{0.471} & \textbf{0.619}& \textbf{0.738} \\
		\bottomrule
	\end{tabular}}
	\caption{Evaluated results on recognition and localization.}
	\label{tab:ResNet50_DenseNet121_CAM}
\end{table}

\subsection{Network Interpretability}
\textbf{Evaluation of Recognition and Localization.} The performance of the CAM generated by the MDM decoder for DProtoNet and the CAM generated by other methods were compared on the eight evaluation indicators, including average drop, average increase, deletion score, insertion score, dice coefficient, IOU, PPV, and sensitivity (Table \ref{tab:ResNet50_DenseNet121_CAM}). DProtoNet with ResNet50 as the backbone improved by 37.1\%, 32.7\%, 26.3\%, 81.3\%, 3.9\%, 4.8\%, 4.3\%, and 1.9\%, respectively, compared with the previous state-of-the-art model; DProtoNet with DenseNet121 as the backbone could improve by 51.3\%, 14.5\%, 8.9\%, 17.9\%, 20.2\%, 29.1\%, 21.6\%, and 20.6\%, respectively, in comparison with the previous state-of-the-art model. CAM generated by the MDM decoder achieved the state of the art in localization and recognition, which had good  interpretability. Based on the result (Figure \ref{fig:deletion_insertion}), the probability curve corresponding to the CAM generated by DProtoNet with ResNet50 as the backbone through the MDM decoder had the sharpest degree of change, implying that DProtoNet can focus on the most meaningful regions for the classification.

\begin{figure}[H]
	\centering
	\includegraphics[width=1.0\linewidth]{{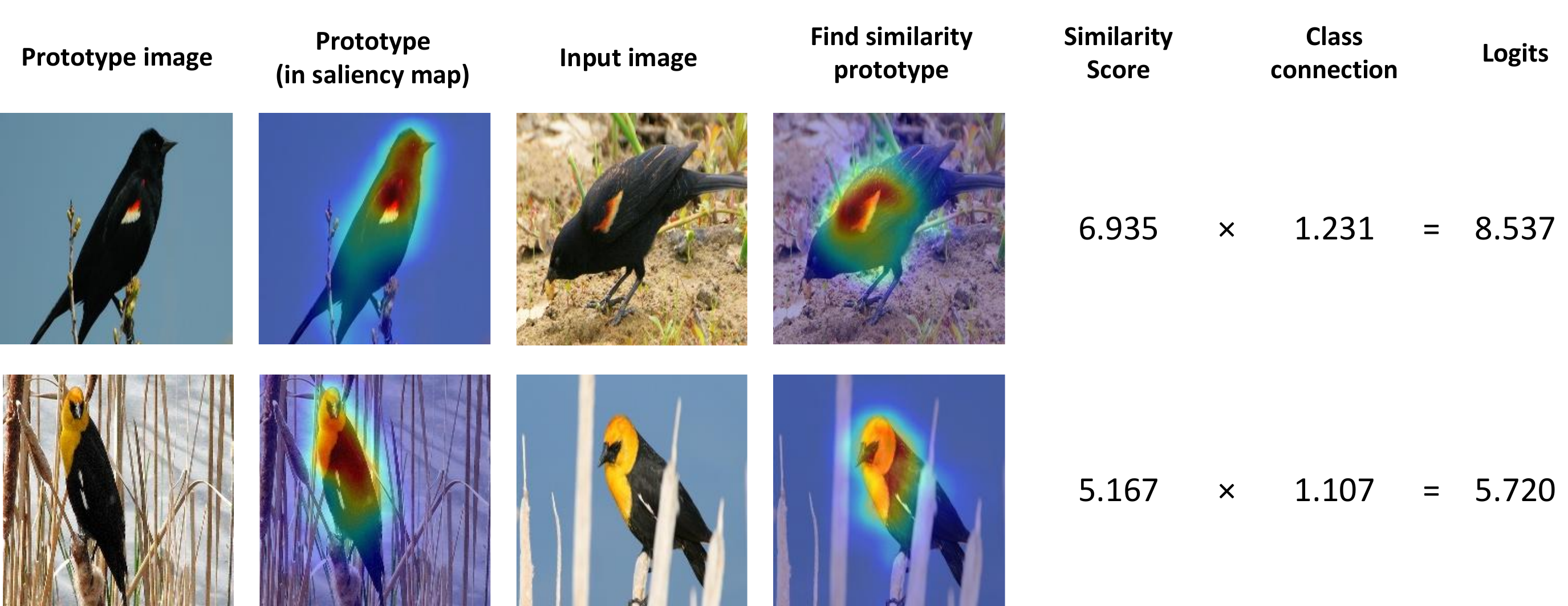}}
	
	\caption{DProtoNet inference process visualization.}
	\label{fig:prototype_inference}
\end{figure}

\begin{figure}[H]
	\centering
	\includegraphics[width=0.8\linewidth]{{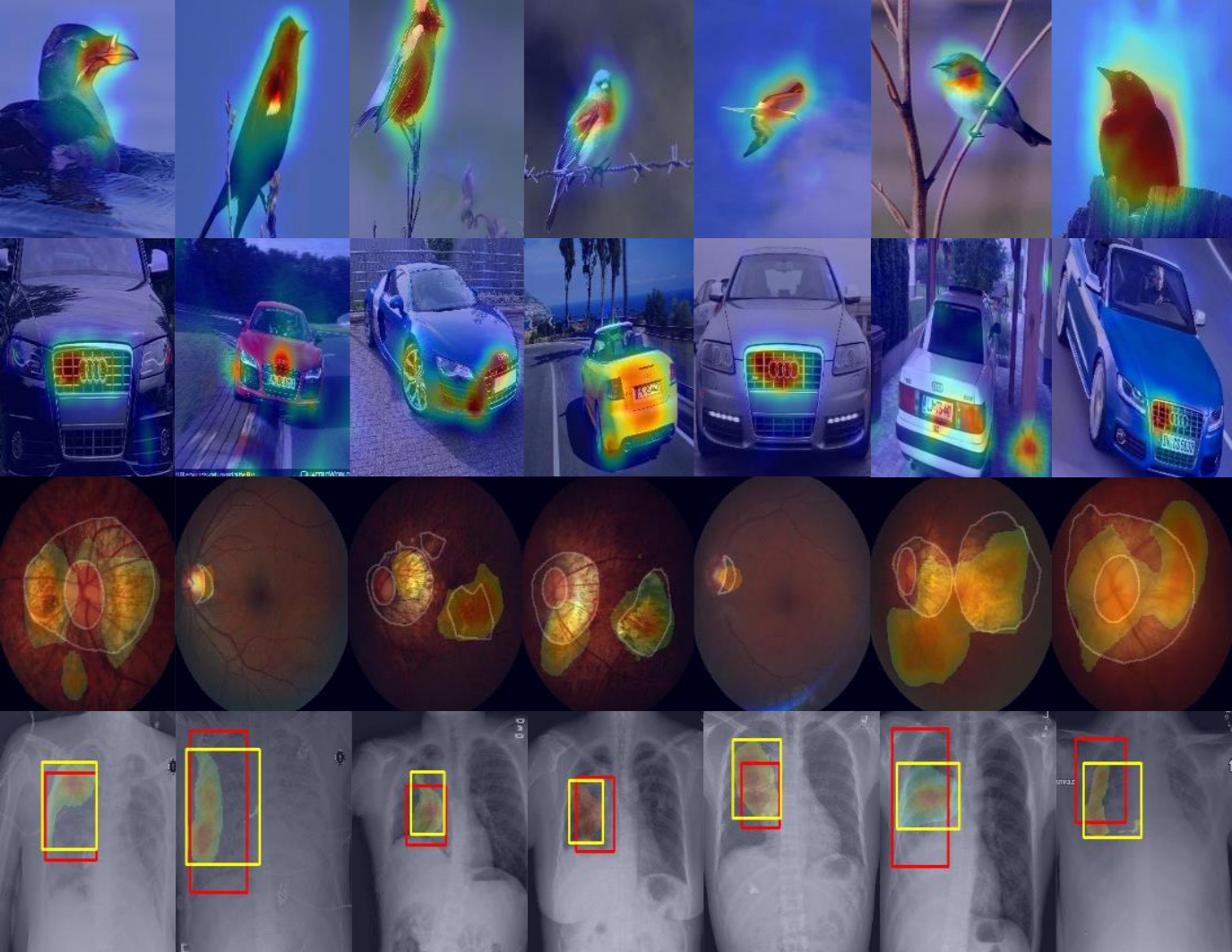}}
	
	\caption{Visualization of decision regions.}
	\label{fig:visualization}
\end{figure}

\textbf{Visualization.} Figure \ref{fig:prototype_inference} shows a visualization of the inference process of DProtoNet, which makes a decision by finding the prototype image that is most similar to the input image and compares the similarity between the input image and the prototype image in the decision region. This is in line with our expectations for the DProtoNet inference process. Figure \ref{fig:visualization} depicts the decision regions found by the MDM decoder when predicting various images. In the visualization, the CAM generated by the MDM decoder is represented by the red to the blue area. Activation and network attention increase from blue to red. The white enclosed area in the fundus retina images \cite{fu2019palm} represents the pathological area that is the ground truth. In the chest X-ray images \cite{gabruseva2020deep}, the red and yellow boxes demonstrate the real lesion area and the bounding box of the lesion area found by the MDM decoder, respectively. According to the finding (Figure \ref{fig:visualization}), the decision region found by the MDM decoder was close to the real decision region, which is similar to the decision basis of human search. Accordingly, the DProtoNet is both interpretable for the inference process and can accurately tell people its decision basis.

\section{Conclusion}
In this paper, an interpretable network (i.e., DProtoNet) was proposed, which can generalize the learning and extraction of prototypes. This network treated the introduced network as a black box, thus it can be universally applied to the existing networks. A general and powerful method (the multiple dynamic masks decoder), which is used to generate saliency maps to represent the decision basis of DProtoNet was proposed in the current paper. The DProtoNet could remove the mutual constraint between accuracy and interpretability and make the network interpretable while preserving the accuracy of the network. Experimental results revealed that the accuracy of our network could outperform the other interpretable neural networks on the four datasets, which is comparable to the performance of the backbone network and has a huge improvement in interpretability. It is hoped that this work paves the way for future research and applications on explainable neural networks.

\bibliography{DProtoNet_pyt}

\appendix

\section{Proof of Multiple Dynamic Masks}
Let: $z$ represents the region in image $x$, and $f_{p}(z)$ denotes the activation of the neural network $f$ at $p$ when the data of the region $z$ is taken as an input. $I(z)=kf_{p}(z)$, where $k$ is a constant greater than zero, $I(z) \in [0,1]$. $I(z)$ is the amount of information that region $z$ contributes to the activation of neural network $f$ at position $p$.

Let: $z$ is all areas of $d_{i}$, and $m$ is the corresponding mask value on it. $z_{1}$ and $z_{2}$ are the two regions of $d_{i}$, $i \in \{1,2,...,N\}$, $m\in [0,1]$, $g$ is the upsampling function.
\begin{equation} \label{s_eq1}
L(m,z) = [f_{p}(z)-f_{p}(mz)]^2 + \eta m
\end{equation}

if $g(z_{1}) \cap g(z_{2})=\varnothing$, then 
\begin{equation} \label{s_eq2}
I(z_{1}+z_{2})=I(z_{1})+I(z_{2})
\end{equation}

if $I(z_{1})<I(z_{2})$, then
\begin{equation} \label{s_eq3}
0 \leq \frac{\partial I(mz_{1})}{\partial m} < \frac{\partial I(mz_{2})}{\partial m}
\end{equation}

Let: $z_{1}$ and $z_{2}$ demonstrate any two disjoint regions of $d_{i}$; $m_{1}$, $m_{2}$ are mask values on $z_{1}$, $z_{2}$. From Equation (\ref{s_eq2}) and Equation (\ref{s_eq3}), the following Equation (\ref{s_eq4}) can be proved, when $L(m, z)$ in Equation (\ref{s_eq1}) achieves the minimum value.
\begin{equation} \label{s_eq4}
(I(z_{1}) - I(z_{2}))(m_{1} - m_{2}) \geq 0
\end{equation}

Reductio ad absurdum. If $L(m, z)$ in Equation (\ref{s_eq1}) has achieved the minimum value, and $\exists z_{1}$, $z_{2}$ satisfy: 
\begin{equation} \label{s_eq5}
(I(z_{1}) - I(z_{2}))(m_{1} - m_{2}) < 0
\end{equation}

Let: $z(d_{i})$ is all areas on $d_{i}$, $z_{0}=z(d_{i})-z_{1}-z_{2}$, and $m_{0}$ is the mask value of $z_{0}$. $g(z_{1}) \cap g(z_{2})=\varnothing$, $g(z_{1}) \cap g(z_{0})=\varnothing$, $g(z_{2}) \cap g(z_{0})=\varnothing$. Due to symmetry, it may be assumed that $I(z_{1}) < I(z_{2})$. From Equations (\ref{s_eq3}) and (\ref{s_eq5}), it can be inferred that $\frac{\partial I(mz_{1})}{\partial m} < \frac{\partial I(mz_{2})}{\partial m}$ and $m_{1} > m_{2}$.

\begin{equation}
\begin{aligned}
& L(m,z) \\
& = L(z_{1}, m_{1}, z_{2}, m_{2}, z_{0},m_{0}) \\
& = [f_{p}(z_{1}+z_{2}+z_{0})-f_{p}(m_{1}z_{1}+m_{2}z_{2} + m_{0}z_{0})]^2 \hspace{32em} \\
&+\eta (m_{1}+m_{2}+m_{0})
\end{aligned}
\end{equation}
\begin{equation}
\begin{aligned}
& L^{'}(m,z) = L(z_{1}, m_{2}, z_{2}, m_{1}, z_{0},m_{0}) \\
& =[f_{p}(z_{1}+z_{2}+z_{0})-f_{p}(m_{2}z_{1}+m_{1}z_{2} + m_{0}z_{0})]^2 \hspace{30.17em}  \\
& +\eta (m_{2}+m_{1}+m_{0}) \hspace{15.59em}
\end{aligned}
\end{equation}
\begin{equation}
\begin{aligned}
& \Delta L \\
& = [2f_{p}(z_{1}+z_{2}+z_{0}) - f_{p}(m_{1}z_{1}+m_{2}z_{2}+m_{0}z_{0}) \\
& - f_{p}(m_{2}z_{1}+m_{1}z_{2}+m_{0}z_{0})] / k \\
& = [2I(z_{1}+z_{2}+z_{0}) - I(m_{1}z_{1}+m_{2}z_{2}+m_{0}z_{0}) \\
& - I(m_{2}z_{1}+m_{1}z_{2}+m_{0}z_{0})] / k^{2} \\
& = \left\{ [2I(z_{1}) + 2I(z_{2}) + 2I(z_{0}) - I(m_{1}z_{1}) - I(m_{2}z_{2}) \right. \hspace{33.37em}\\
& - I(m_{0}z_{0}) - I(m_{2}z_{1}) - I(m_{1}z_{2}) - I(m_{0}z_{0})]\left.\right\}/ k^{2} \\
& = \left\{[I(z_{1}) - I(m_{1}z_{1})] + [I(z_{2}) - I(m_{1}z_{2})] \right.\\
& + [I(z_{1}) - I(m_{2}z_{1})] + [I(z_{2}) - I(m_{2}z_{2})] \\
& + 2[I(z_{0}) - I(m_{0}z_{0})]\left.\right\}/k^{2} \\
& = \int_{m_{1}}^{1}\frac{\partial I(mz_{1})}{k^{2}\partial m}dm + \int_{m_{1}}^{1}\frac{\partial I(mz_{2})}{k^{2}\partial m}dm \\
& + \int_{m_{2}}^{1}\frac{\partial I(mz_{1})}{k^{2}\partial m}dm + \int_{m_{2}}^{1}\frac{\partial I(mz_{2})}{k^{2}\partial m}dm \\
& + 2\int_{m_{0}}^{1}\frac{\partial I(mz_{0})}{k^{2}\partial m}dm > 0 
\end{aligned}
\end{equation}
\begin{equation}
\begin{aligned}
& L^{'}(m,z)-L(m,z) \\
& = k \Delta L [f_{p}(m_{1}z_{1}+m_{2}z_{2}+ m_{0}z_{0}) \\
& - f_{p}(m_{2}z_{1} +m_{1}z_{2} + m_{0}z_{0})] \\
& = \Delta L [I(m_{1}z_{1}+m_{2}z_{2}+m_{0}z_{0}) \\
& - I(m_{2}z_{1} +m_{1}z_{2} + m_{0}z_{0})] \\
& = \Delta L [I(m_{1}z_{1})+I(m_{2}z_{2})+I(m_{0}z_{0}) \hspace{39.37em} \\
&  - I(m_{2}z_{1}) - I(m_{1}z_{2}) - I(m_{0}z_{0})] \\
& = \Delta L \left\{[I(m_{1}z_{1}) - I(m_{2}z_{1})] - [I(m_{1}z_{2}) - I(m_{2}z_{2})]   \right\} \\
& = \Delta L \int_{m_{2}}^{m_{1}}[\frac{\partial I(mz_{1})}{\partial m} - \frac{\partial I(mz_{2})}{\partial m}]dm < 0
\end{aligned}
\end{equation}
$L^{'}(m,z)<L(m,z)$, which contradicts that $L$ has achieved a minimum. Therefore, Equation (\ref{s_eq4}) holds.

\section{Explanation of Prototype Expressiveness}
Considering that feature maps $A \in R^{H_{1} \times W_{1} \times D_{1}}$. We represent feature maps by matrix:
\begin{equation}
\begin{aligned}
A=[a_{ij}]_{H_{1} \times W_{1}}, \  a_{ij} = [a^{1}_{ij}, a^{2}_{ij}, ..., a^{D_{1}}_{ij}]^{T}
\end{aligned}
\end{equation}

Previous prototyped-based models \cite{chen2019looks,singh2021these, singh2021interpretable, kim2021xprotonet, hase2019interpretable, li2018deep} can be divided into three types of prototype extraction methods represented by ProtoPNet \cite{chen2019looks}, Gen-ProtoPNet \cite{singh2021interpretable}, and XProtoNet \cite{kim2021xprotonet}. Note that $Z_{P}$, $Z_{G}$, $Z_{X}$, and $Z_{D}$ respectively are the sets of all prototypes that can be extracted by ProtoPNet, Gen-ProtoPNet, XProtoNet, and DProtoNet. From these models and DProtoNet, the sets of $Z_{P}, Z_{G}, Z_{X}$, and $Z_{D}$ can be expressed as follows, respectively:
\begin{equation}
\begin{aligned}
Z_{P}=\{[a_{i,j}]_{1 \times 1} |i \in \{1,2,...,H_{1}\}, j \in \{1,2,...,W_{1}\} \} \hspace{36em}
\end{aligned}
\end{equation}
\begin{equation}
\begin{aligned}
Z_{G} = & \{[a_{i+u,j+v}]_{h \times w} |i \in \{1,2,...,h\}, \\
& j \in \{1,2,...,w\}, u \in \{0,1,..., H_{1} - h \}, \\
& v \in \{0,1,..., W_{1} - w \}, 1< hw < H_{1}W_{1} \} \hspace{36em}
\end{aligned}
\end{equation}
\begin{equation}
\begin{aligned}
Z_{X}=\{B*A |B=[b_{i,j}]_{H_{1} \times W_{1}}, b_{i,j} \in [0, 1] \} \hspace{39em}
\end{aligned}
\end{equation}
\begin{equation}
\begin{aligned}
Z_{D} = & \{C*A | C=[c_{i,j}]_{H_{1} \times W_{1}}, \hspace{37em} \\
& c_{i,j} = [c_{i,j}^{1}, c_{i,j}^{2},..., c_{i,j}^{D_{1}}]^{T},\\
& c_{i,j}^{d} \in [0, 1], d \in \{1, 2, ..., D_{1}\}	\}
\end{aligned}
\end{equation}
\begin{equation}
\begin{aligned}
Z^{1}_{D} = & \{C^{1}*A | C^{1}=[c_{i,j}]_{H_{1} \times W_{1}}, c_{u,v} = [1, 1, ..., 1]^{T}, \hspace{36em} \\
& if \ m \neq u \ or \  n \neq v, then \ c_{m,n} = [0, 0, ..., 0]^{T}, \\
& u \in \{1,2,...,H_{1}\}, v \in \{1,2,...,W_{1}\}	\}
\end{aligned}
\end{equation}
\begin{equation}
\begin{aligned}
Z^{2}_{D} = & \{C^{2}*A | C^{2}=[c_{ij}]_{H_{1} \times W_{1}}, \hspace{34em} \\
& c_{u+r,v+s} = [1, 1, ..., 1]^{T}, 1 \leq r \leq h, 1 \leq s \leq w, \\ 
& 0 \leq u \leq H_{1} - h , 0 \leq v \leq W_{1} - w, \\
& if \ m \leq u \ or \ m > u+h \  or \ n \leq v \ or \ n > v+w, \\
& \  then \ c_{m,n} = [0, 0, ..., 0]^{T}, 1< hw < H_{1}W_{1} \}
\end{aligned}
\end{equation}
\begin{equation}
\begin{aligned}
Z^{3}_{D} = & \{C^{3}*A | C^{3}=[c_{i,j}]_{H_{1} \times W_{1}}, \hspace{37em} \\
& c_{i,j} = [c_{i,j}^{1}, c_{i,j}^{2},..., c_{i,j}^{D_{1}}]^{T},c_{i,j}^{d} = \epsilon_{i,j}, \\
& \epsilon_{i,j} \in [0, 1], d \in \{1, 2, ..., D_{1}\}	\}
\end{aligned}
\end{equation}
where $*$ is the Hadamard product.

$C^{i} \subseteq C$, $Z^{i}_{D} \subseteq Z_{D}$ $(i=1,2,3)$. $Z^{1}_{D}$ and $Z^{2}_{D}$ can generate $Z_{P}$ and $Z_{G}$ respectively by removing several 0 matrices, and $Z^{3}_{D} = Z_{X}$. Therefore, $Z_{D}$ can generate $Z_{P}$, $Z_{G}$, $Z_{X}$. The set of prototypes generated by DProtoNet includes the sets of prototypes generated by previous models.

The number of elements of the sets $Z_{P}$, $Z_{G}$, $Z_{X}$, and $Z_{D}$ are as follows, respectively:

\begin{algorithm}[tb]
	\caption{Multiple Dynamic Masks Decoder}
	\label{alg:algorithm1}
	\textbf{Input}: Image $X_{0}$, Neural Network $f(x)$, Activation Position $p$, Upsampling Function $g(x)$, Loss Function $L$.\\
	\textbf{Output}: Heatmap $M^{h}$, Binary Mask $M^{b}$, Heatmap Image $M^{h}_{MDM}$, Binary Mask Image $M^{b}_{MDM}$.\\
	\textbf{Parameter}: Weights $\{\lambda_{i}\}^{N}_{i=1}$, Mask Feature Vectors $\{d_{i}\}^{N}_{i=1}$,
	Training Epochs $C$, Learning Rate $\eta$, Threshold $\gamma$, Mix Weights $\alpha$, $\beta$.
	\begin{algorithmic}[1] 
		\STATE $A^{p} \gets f_{p}(X_{0})$
		\STATE \textbf{for} $i=1$ \textbf{to} $N$ \textbf{do}
		\STATE \;\;\,\,\,\, Initialize $d_{i}$ each element is 0.5
		\STATE \;\;\,\,\,\, \textbf{for} $j=1$ \textbf{to} $C$ \textbf{do}
		\STATE \;\;\,\,\,\,\;\;\,\,\,\,\, $M_{i} \gets g(d_{i})$
		\STATE \;\;\,\,\,\,\;\;\,\,\,\,\, $A^{p}_{i} \gets f_{p}(M_{i} \cdot X_{0})$
		\STATE \;\;\,\,\,\,\;\;\,\,\,\,\, $L_{c} \gets L(A^{p}, A^{p}_{i})$
		\STATE \;\;\,\,\,\,\;\;\,\,\,\,\, $L_{d} \gets ||d_{i}||_{1}$
		\STATE \;\;\,\,\,\,\;\;\,\,\,\,\, $L_{t} \gets L_{c} + \lambda_{i}L_{d}$
		\STATE \;\;\,\,\,\,\;\;\,\,\,\,\, $\theta_{d_{i}} \gets \theta_{d_{i}} - \eta \frac{\partial L_{t}}{\partial \theta_{d_{i}}}$
		\STATE \;\;\,\,\,\, \textbf{end for}
		\STATE \textbf{end for}
		\STATE Initialize $M^{F}$ to zero mask
		\STATE \textbf{for} $i=1$ \textbf{to} $N$ \textbf{do}
		\STATE \;\;\,\,\,\,\, $M^{F} \gets M^{F} + g(d_{i})$
		\STATE \textbf{end for}
		\STATE $M^{b} = \{M^{F} \geq \gamma\}$
		\STATE $M^{h} = M^{b} \cdot M^{F}$
		\STATE Normalize $M^{h}$
		\STATE $M_{MDM}^{h}=\alpha X_{0} + \beta M^{h}$
		\STATE $M^{b}_{MDM} = M^{b} \cdot X_{0}$
		\STATE \textbf{return} $M^{h}$, $M^{b}$, $M^{h}_{MDM}$, $M^{b}_{MDM}$
	\end{algorithmic}
\end{algorithm}

\begin{equation} \label{zp}
\begin{aligned}
|Z_{P}|= H_{1}W_{1} \hspace{36.86em}
\end{aligned}
\end{equation}
\begin{equation} \label{zg}
\begin{aligned}
|Z_{G}| = \frac{H_{1}W_{1}(H_{1}W_{1} + H_{1} + W_{1} -3)}{4}-1 \hspace{22.86em}
\end{aligned}
\end{equation}
\begin{equation} \label{zx}
\begin{aligned}
|Z_{X}|= \prod_{1 \leq i \leq H_{1}, 1 \leq j \leq W_{1}} n(\epsilon_{i,j}) \hspace{28.86em}
\end{aligned}
\end{equation}
\begin{equation} \label{zd}
\begin{aligned}
|Z_{D}|= \prod_{1 \leq i \leq H_{1}, 1 \leq j \leq W_{1}, 1 \leq d \leq D_{1}} n(\epsilon^{d}_{i,j}) \hspace{18.86em}
\end{aligned}
\end{equation}
where $\epsilon_{i,j}, \epsilon^{d}_{i,j} \in [0,1]$, $n(x)$ represents the number of different elements that variable $x$ can produce. From Equations (\ref{zp}), (\ref{zg}), (\ref{zx}), and (\ref{zd}), the following Equation (\ref{ieq}) can be achieved:
\begin{equation} \label{ieq}
\begin{aligned}
|Z_{P}| < |Z_{G}| < H_{1}^{2}W_{1}^{2} \ll |Z_{X}| \ll |Z_{D}|
\end{aligned}
\end{equation}

The expression ability of the prototype generated by DProtoNet is far greater than that of the previous models.

\begin{algorithm}[tb]
	\caption{Parameters Update in DProtoNet Training}
	\label{alg:algorithm2}
	\textbf{Input}: Shaping Layer Parameters $\theta_{a}$, Backbone Layer Parameters $\theta_{b}$, Prototype Layer Parameters $\theta_{g}$, Fully Connected Layer Parameters $\theta_{h}$, Training Epochs $C$, Jointly Stage $C_{j}$, Push Stages $C_{p}$, Iterations $N$. \\
	\textbf{Output}: $\theta_{a}$, $\theta_{b}$, $\theta_{g}$, $\theta_{h}.$ 
	\begin{algorithmic}[1] 
		\STATE \textbf{for} $i=1$ \textbf{to} $C$ \textbf{do}
		\STATE \;\;\,\,\,\,\, Calculate network loss
		\STATE \;\;\,\,\,\,\, \textbf{if} $i < C_{j}$ \textbf{then} 
		\STATE \;\;\,\,\,\,\;\;\,\,\, Update $\theta_{a}, \theta_{g}, \theta_{h}$ with SGD
		\STATE \;\;\,\,\,\,\, \textbf{else}
		\STATE \;\;\,\,\,\,\;\;\,\,\, Update $\theta_{a}, \theta_{b}, \theta_{g}, \theta_{h}$ with SGD
		\STATE \;\;\,\,\,\,\;\;\,\,\, \textbf{if} $ i$ in $C_{p}$ \textbf{then}
		\STATE \;\;\,\,\,\,\;\;\,\,\,\,\,\,\,\, Update prototypes
		\STATE \;\;\,\,\,\,\;\;\,\,\,\,\,\,\,\, \textbf{for} $t=1$ \textbf{to} $N$ \textbf{do}
		\STATE \;\;\,\,\,\,\;\;\,\,\,\,\,\,\,\,\,\,\,\,\,\,\,\,\, Calculate network loss
		\STATE \;\;\,\,\,\,\;\;\,\,\,\,\,\,\,\,\,\,\,\,\,\,\,\,\, Update $\theta_{h}$ with SGD
		\STATE \;\;\,\,\,\,\;\;\,\,\,\,\,\,\,\, \textbf{end for}
		\STATE \;\;\,\,\,\,\;\;\,\,\, \textbf{end if}
		\STATE \;\;\,\,\,\,\, \textbf{end if}
		\STATE \textbf{end for}
		\STATE \textbf{return} $\theta_{a}$, $\theta_{b}$, $\theta_{g}$, $\theta_{h}$
	\end{algorithmic}
\end{algorithm}

\section{Additional Experimental Results}
\begin{figure}[H]
	\centering
	\includegraphics[width=0.7\linewidth]{{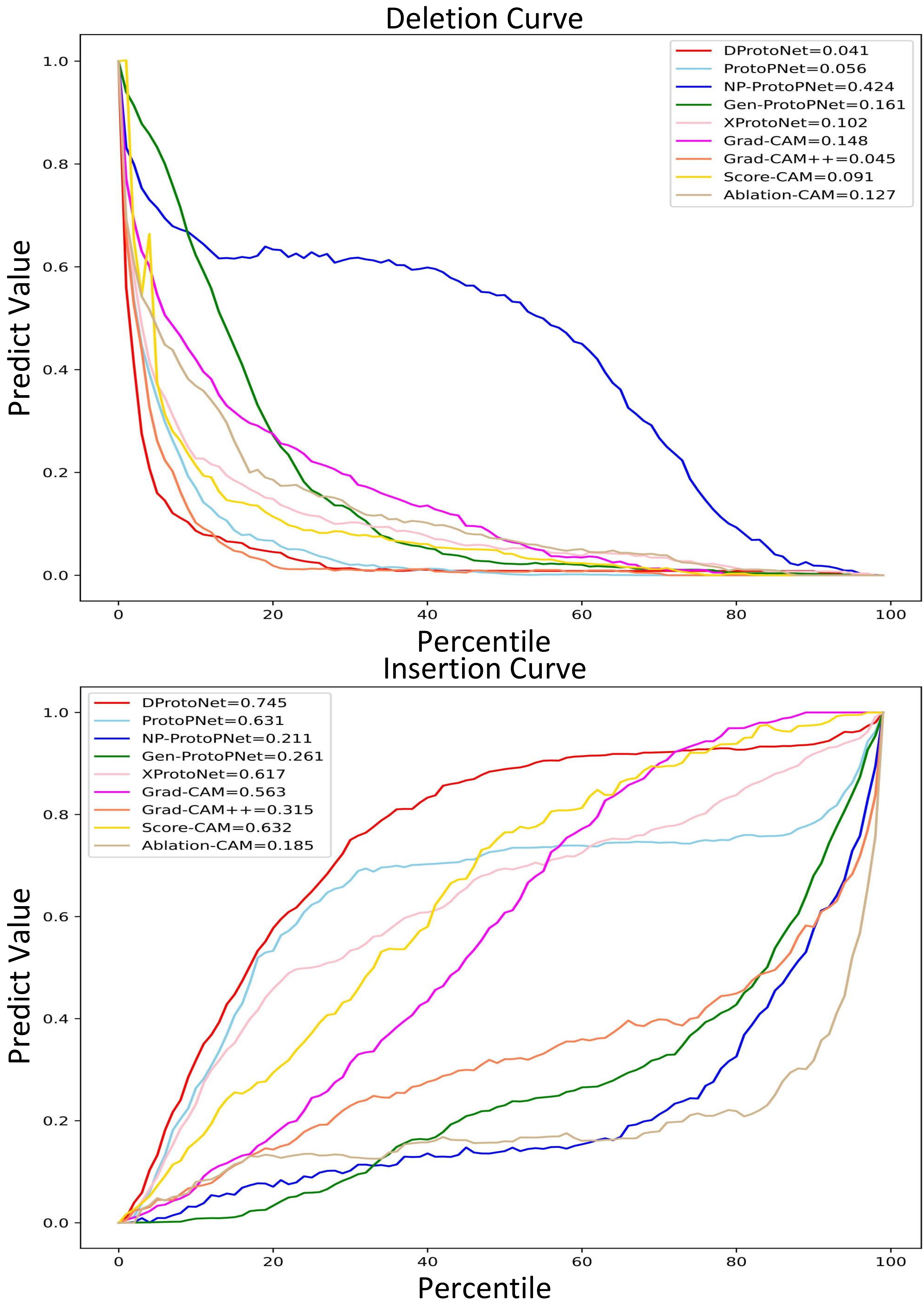}}
	
	\caption{Deletion and insertion curves of the multiple methods.}
	\label{del_ins_densenet121}
\end{figure}

In order to further evaluate the recognition \cite{wang2020score} ability of the class activation maps (CAM) \cite{zhou2016learning} generated by DProtoNet, we compared the deletion and insertion scores \cite{wang2020score} of multiple methods on DenseNet121 \cite{huang2017densely} as the backbone. Based on the result (Figure \ref{del_ins_densenet121}), DProtoNet achieved the state of the art on the deletion and insertion scores. The CAM generated by DProtoNet with DenseNet121 as the backbone achieved the most drastically changing probability curve, implying that DProtoNet can focus on the most meaningful regions for the classification.

In order to compare the localization \cite{wang2020score} ability of the CAM generated by multiple methods in detail, the binary mask thresholds were set to traverse the top 1\% to 99\% of the activation values of the CAM to exam the dice coefficient, IOU, PPV, and sensitivity \cite{laradji2021weakly}.

\begin{figure}[H]
	\centering
	\includegraphics[width=0.9\textwidth]{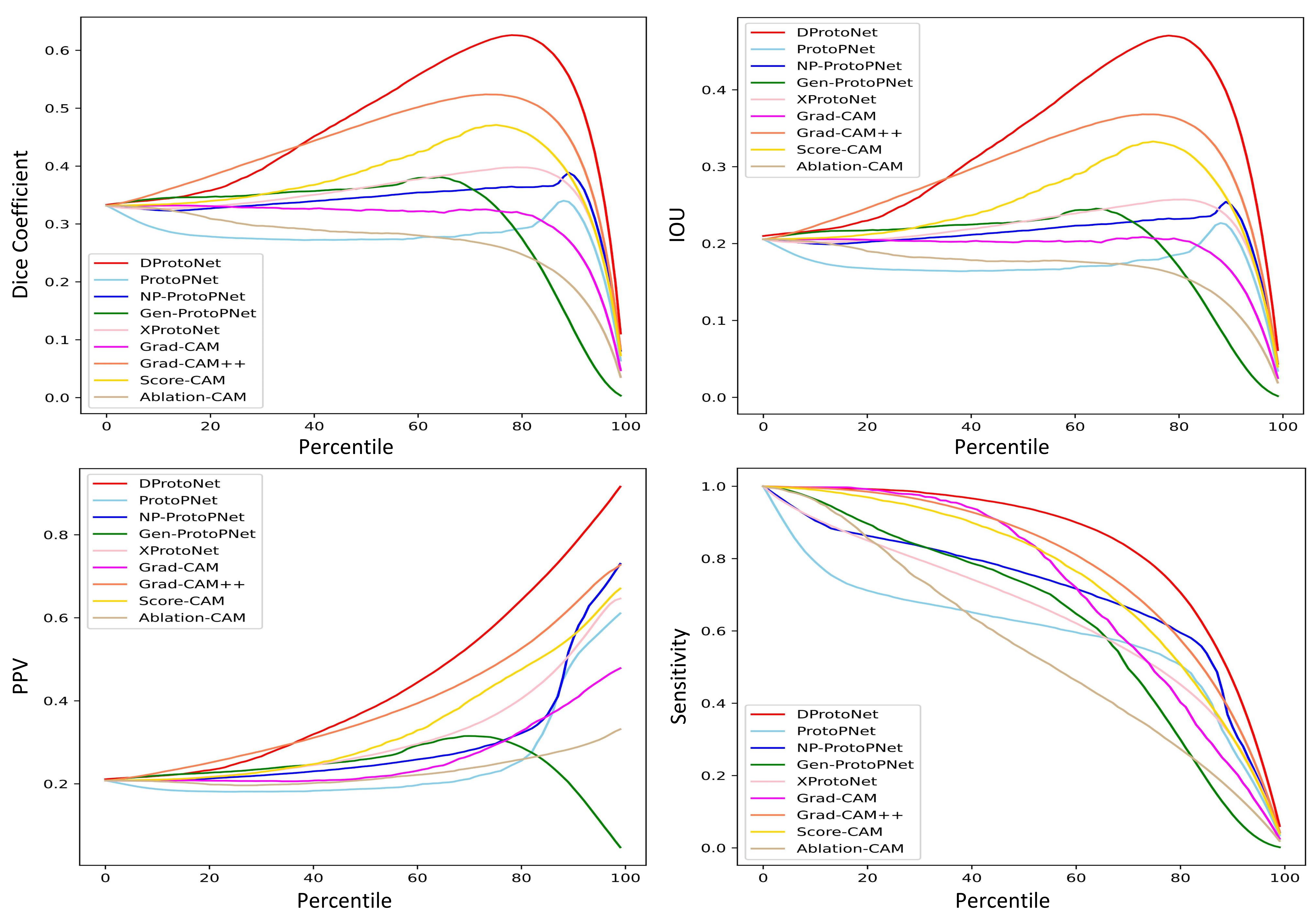}

	\caption{Dice Coefficient, IOU, PPV, and Sensitivity curves calculated by nine methods, with ResNet50 as the backbone.}
	\label{disp_resnet50}
\end{figure}

\begin{figure}[H]
	\centering
	\includegraphics[width=0.9\textwidth]{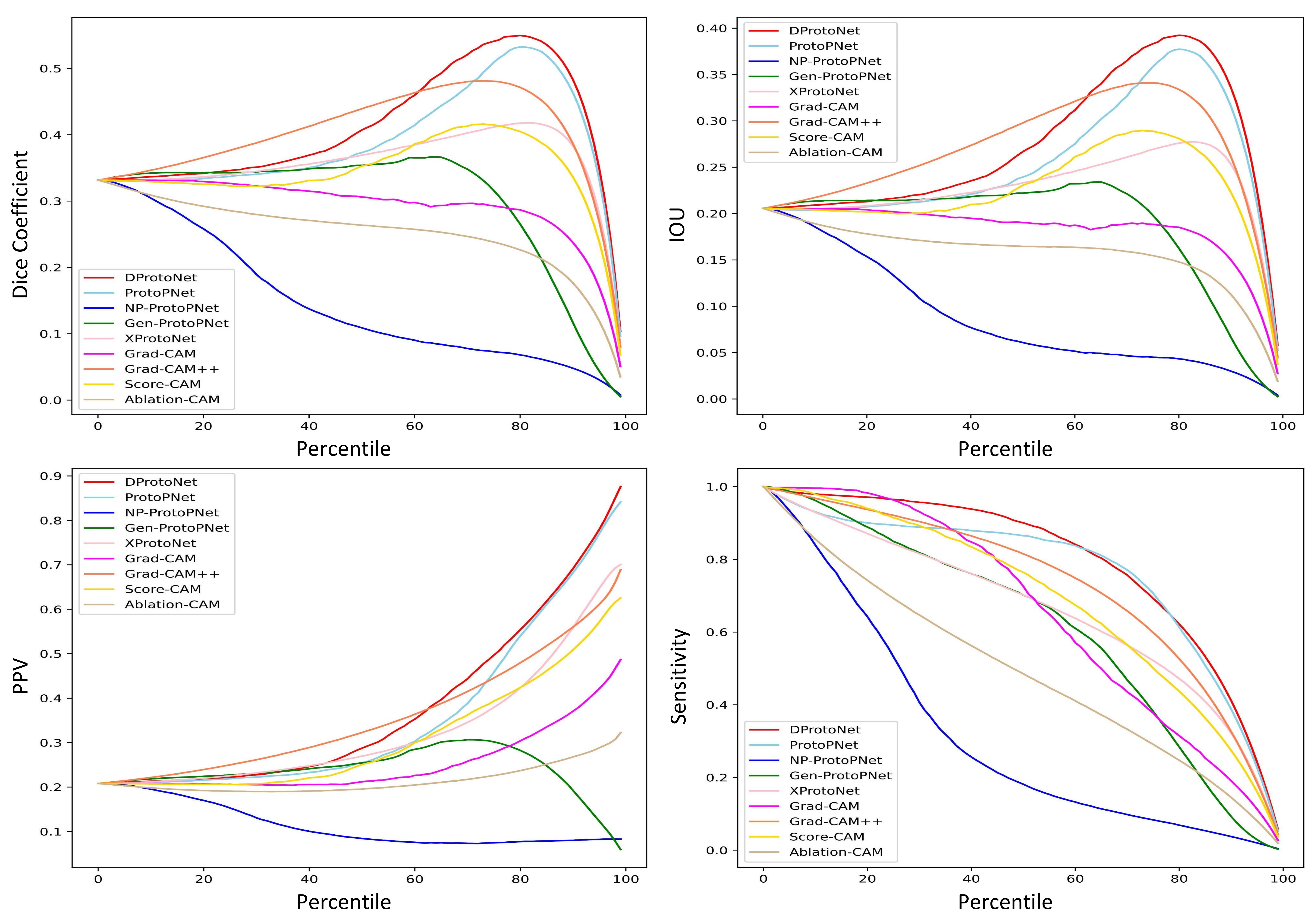}

	\caption{Dice Coefficient, IOU, PPV, and Sensitivity curves calculated by nine methods, with DenseNet121 as the backbone.}
	\label{disp_densenet121}
\end{figure}

Figures \ref{disp_resnet50} and \ref{disp_densenet121} show the results of the foreground image and binary mask corresponding to the CAM generated by various methods on each threshold in the above-mentioned evaluation indicators on the birds dataset \cite{wah2011caltech}. The results show that the top 1\% to 60\% and 1\% to 35\% of the pixels in the activation degree of the CAM generated by DProtoNet with ResNet50 and DenseNet121 as the backbone achieved the state of the art localization ability, respectively.

\section{Pseudo Code}
To describe MDM decoder and DProtoNet in detail, the pseudo codes of the workflow of MDM decoder and training scheme of DProtoNet as shown in Algorithms \ref{alg:algorithm1} and \ref{alg:algorithm2}.

\section{Visualization}
\begin{figure*} 
	\centering
	\includegraphics[width=0.9\textwidth]{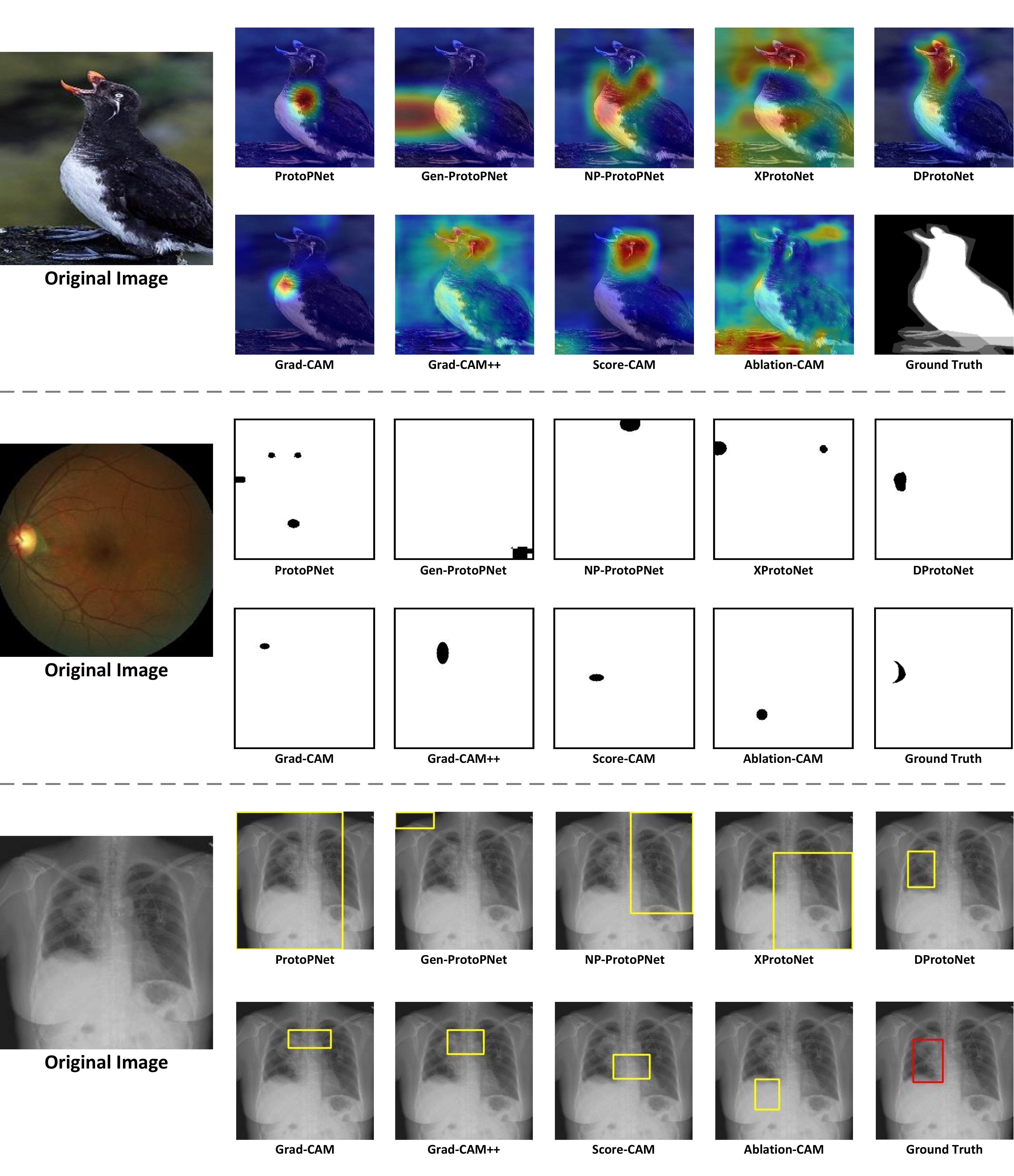}
	\hfill
	\caption{Visual comparison results on the bird, fundus retina, and chest X-ray images, respectively.}
	\label{com_v_img1}
\end{figure*}

\begin{figure*} 
	\centering
	\includegraphics[width=0.9\textwidth]{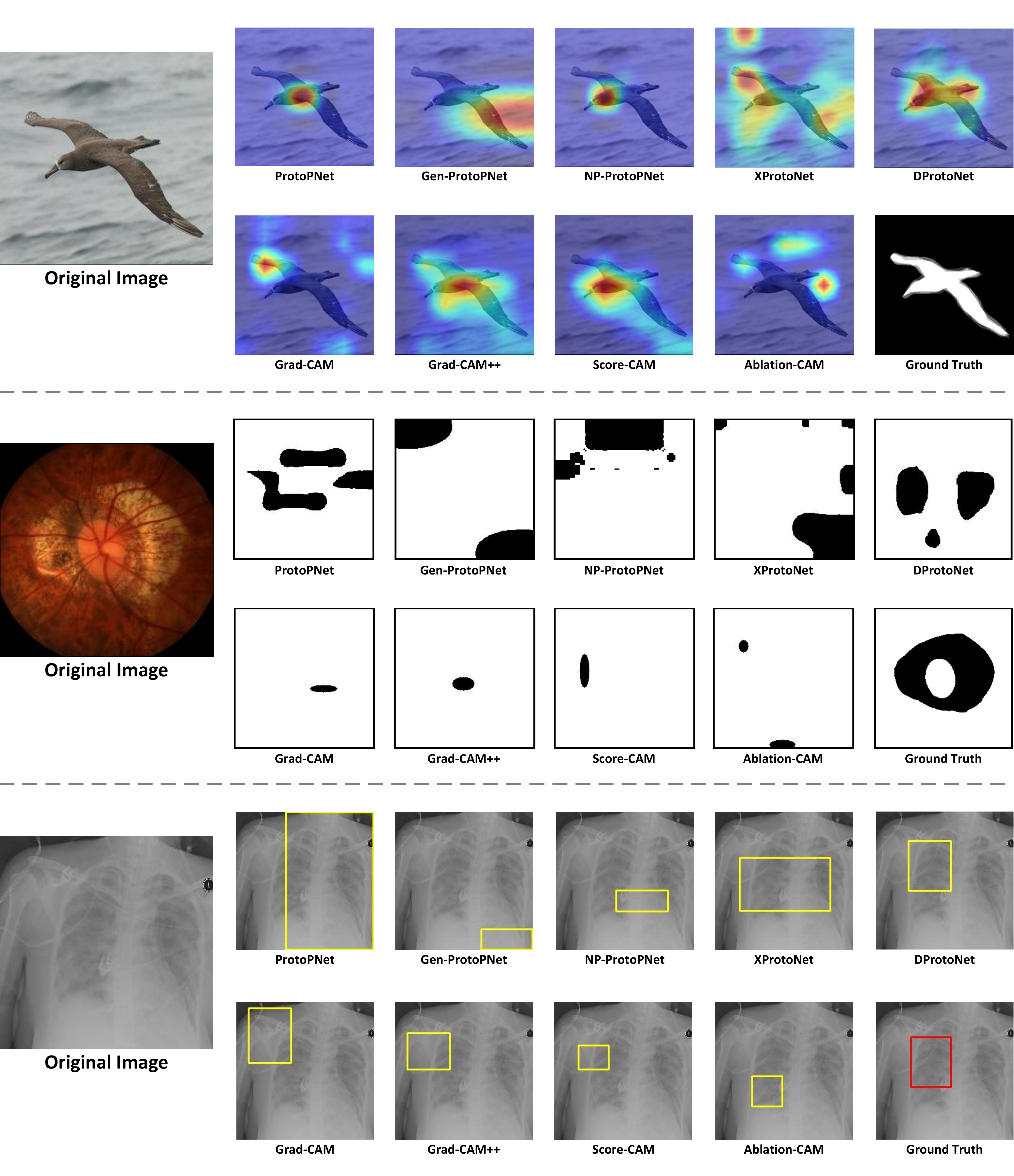}
	\hfill
	\caption{Visual comparison results on the bird, fundus retina, and chest X-ray images, respectively.}
	\label{com_v_img2}
\end{figure*}

\begin{figure*} 
	\centering
	\includegraphics[width=0.9\textwidth]{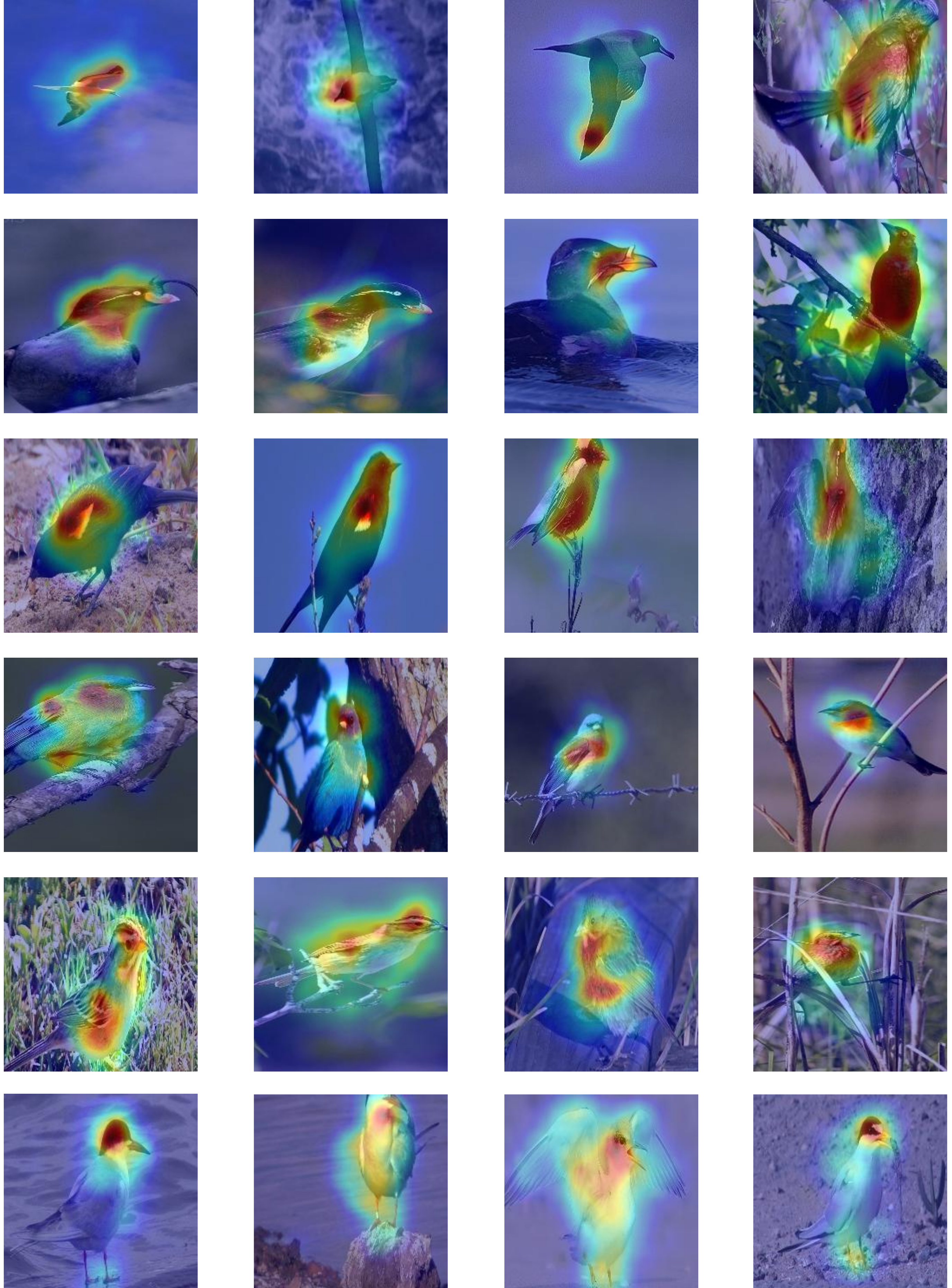}
	\hfill
	\caption{Visualization of DProtoNet decision regions in bird images. From blue to red, the activation degree increase.}
	\label{v_img1}
\end{figure*}

\begin{figure*} 
	\centering
	\includegraphics[width=0.9\textwidth]{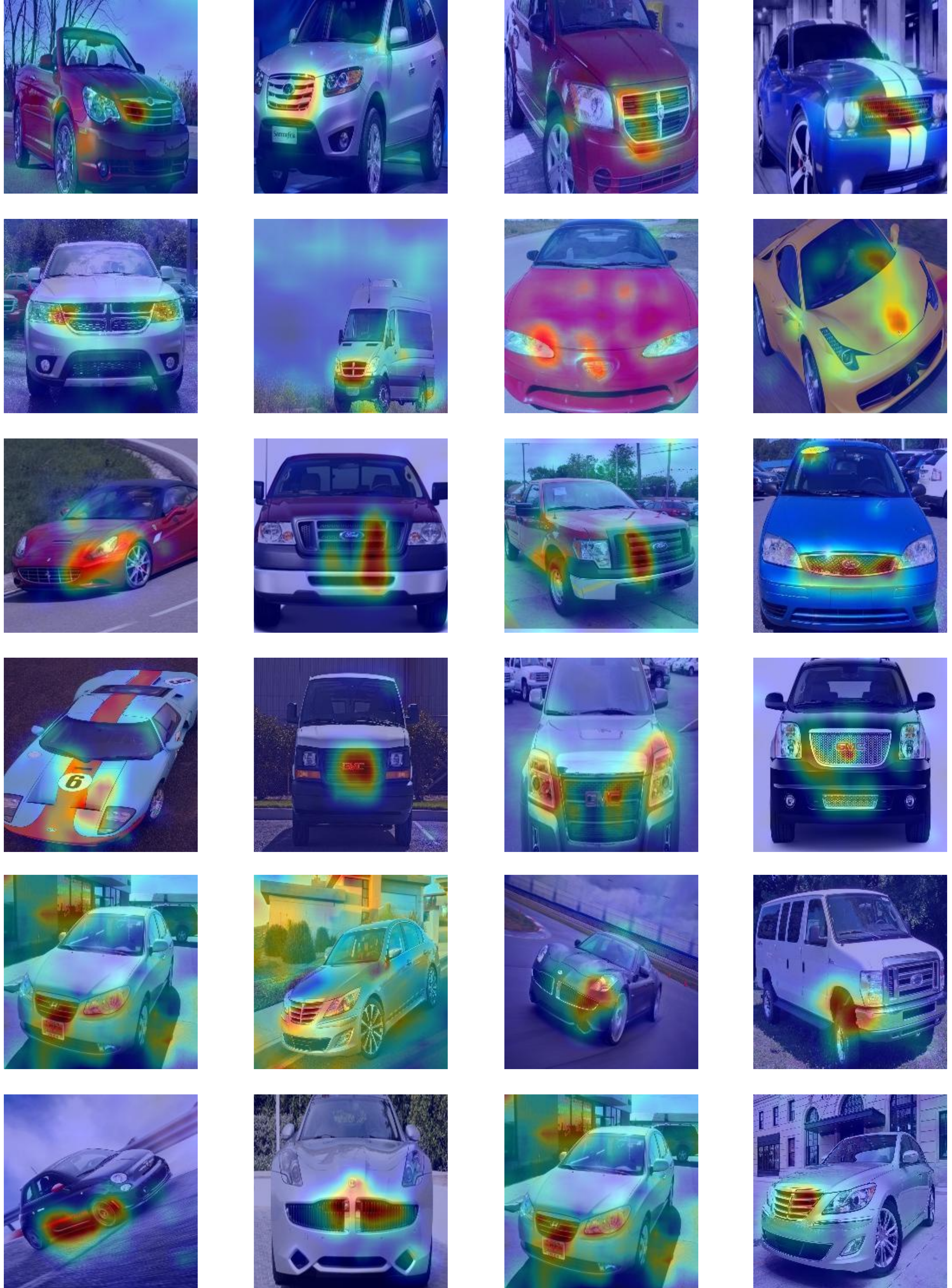}
	\hfill
	\caption{Visualization of DProtoNet decision regions in car images. From blue to red, the activation degree increase.}
	\label{v_img2}
\end{figure*}

\begin{figure*}
	\centering
	\includegraphics[width=0.9\textwidth]{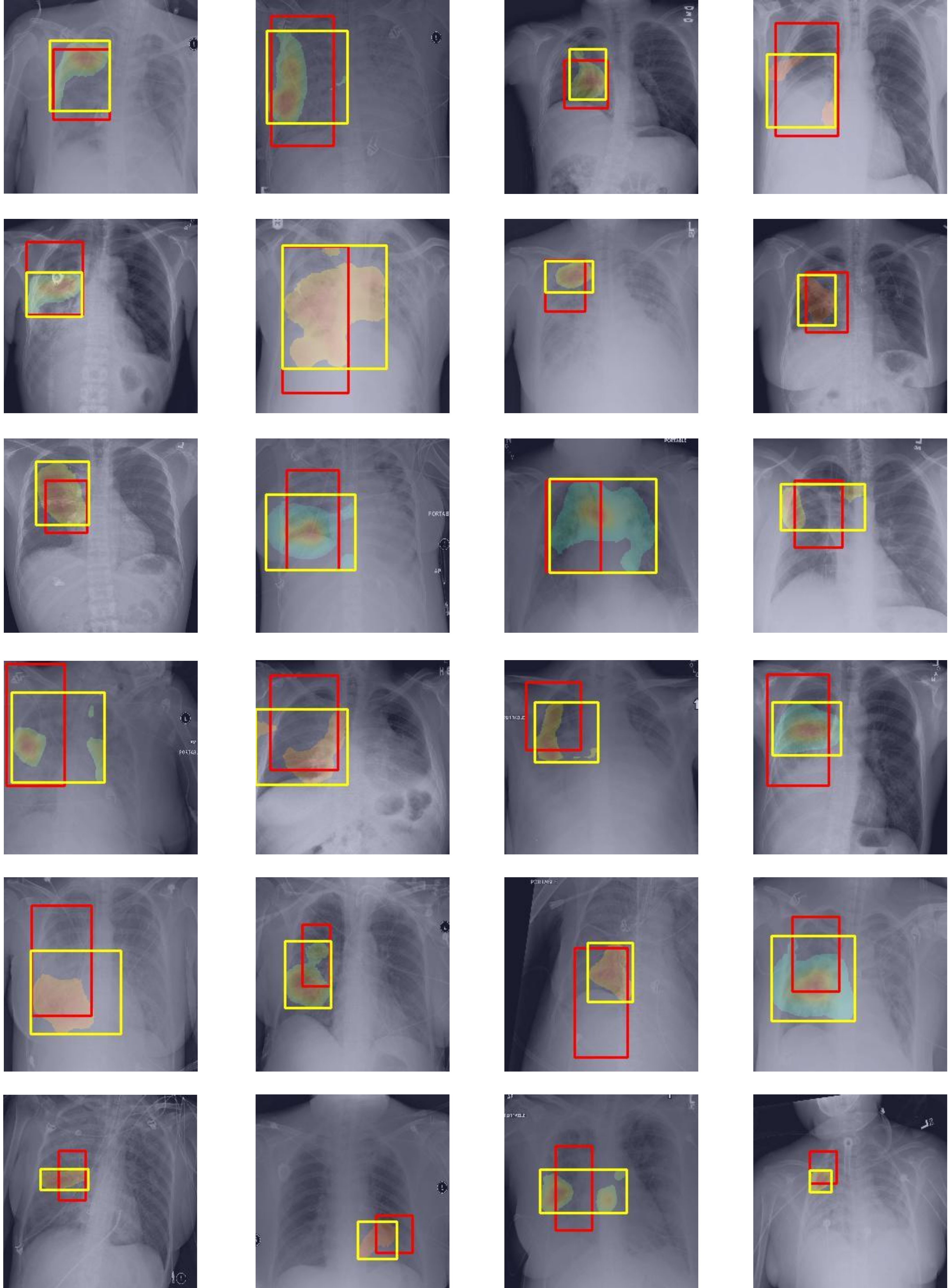}
	\hfill
	\caption{Visualization of DProtoNet decision regions in chest X-ray images. The red and yellow boxes demonstrate the real lesion area and the lesion area found by the DProtoNet, respectively.}
	\label{v_img3}
\end{figure*}

\begin{figure*} 
	\centering
	\includegraphics[width=0.9\textwidth]{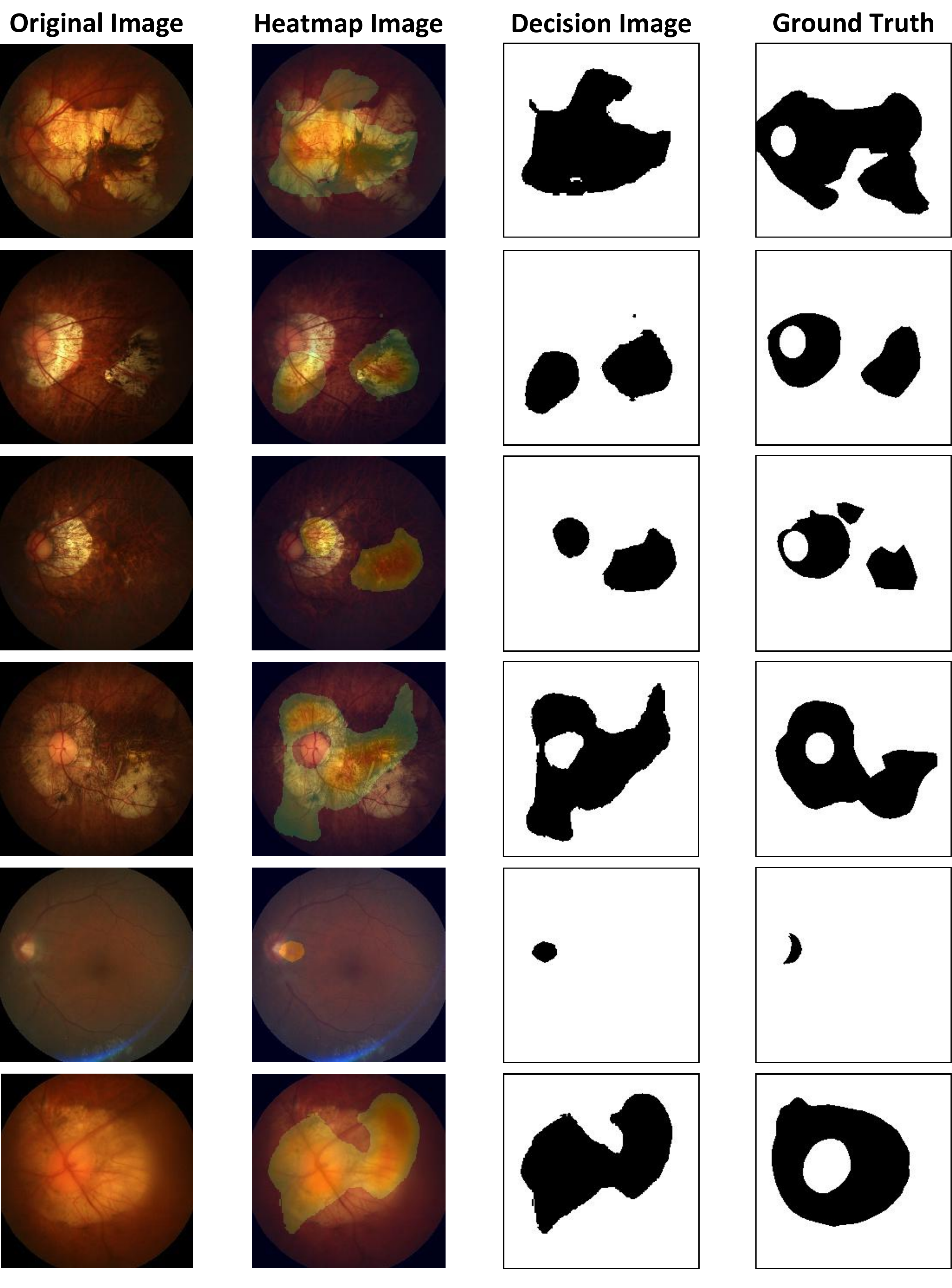}
	\hfill
	\caption{Visualization of DProtoNet decision regions in fundus retina images.}
	\label{v_img4}
\end{figure*}

In this section, we provide more visualization of decision regions for DProtoNet inference to verify the effectiveness of it. We compare the visualization of the decision regions of many methods on bird \cite{wah2011caltech}, fundus retina \cite{fu2019palm}, and chest X-ray images \cite{gabruseva2020deep}, as shown in Figures \ref{com_v_img1} and \ref{com_v_img2}. We show the visualization of DProtoNet's classification decision basis for bird \cite{wah2011caltech}, car \cite{krause20133d}, fundus retina \cite{fu2019palm}, and chest X-ray images \cite{gabruseva2020deep} in Figures \ref{v_img1}, \ref{v_img2}, \ref{v_img3}, and \ref{v_img4}.

\end{document}